\documentclass{article}
\usepackage{ijcai24}

\usepackage{times}
\usepackage{soul}
\usepackage{url}
\usepackage[hidelinks]{hyperref}
\usepackage[utf8]{inputenc}
\usepackage[small]{caption}
\usepackage{graphicx}
\usepackage{amsmath}
\usepackage{amsthm}
\usepackage{booktabs}
\usepackage{algorithm}
\usepackage{algorithmic}
\usepackage[switch]{lineno}
\usepackage{subfigure}

\usepackage{algorithm}
\usepackage{algorithmic}
\usepackage{comment}
\usepackage{graphicx}
\usepackage{amsmath}
\usepackage{amssymb}
\usepackage{booktabs}
\usepackage{enumitem}
\usepackage{xcolor}
\usepackage{url}
\newcommand\crule[3][black]{\textcolor{#1}{\rule{#2}{#3}}}
\usepackage{booktabs, multirow} 
\definecolor{roadcolor}{RGB}{234,51,246}
\definecolor{sidewalkcolor}{RGB}{68,8,72}
\definecolor{parkingcolor}{RGB}{241,156,249}
\definecolor{othergroundcolor}{RGB}{160,32,76}
\definecolor{buildingcolor}{RGB}{246,202,69}
\definecolor{carcolor}{RGB}{111,149,238}
\definecolor{truckcolor}{RGB}{74,32,172}
\definecolor{bicyclecolor}{RGB}{136,227,242}
\definecolor{motorcyclecolor}{RGB}{37,59,146}
\definecolor{othervehiclecolor}{RGB}{96,81,242}
\definecolor{vegetationcolor}{RGB}{79, 173, 50}
\definecolor{trunkcolor}{RGB}{126, 65, 22}
\definecolor{terraincolor}{RGB}{171, 238, 105}
\definecolor{personcolor}{RGB}{234, 60, 49}
\definecolor{bicyclistcolor}{RGB}{234, 66, 195}
\definecolor{motorcyclistcolor}{RGB}{138, 42, 90}
\definecolor{fencecolor}{RGB}{238, 128, 69}
\definecolor{polecolor}{RGB}{252, 241, 161}
\definecolor{trafficsigncolor}{RGB}{233, 51, 35}
\definecolor{color1}{RGB}{176, 36, 24}
\definecolor{color2}{RGB}{119,185,0}
\definecolor{color3}{RGB}{0, 0, 200}
\definecolor{colorofteaser}{RGB}{176, 36, 24}
\definecolor{color4}{RGB}{0, 0, 0}

\newcommand{\cmark}{\ding{51}}%
\usepackage{pifont}

\newcommand{\tbb}[1]{\textbf{\textcolor{color4}{#1}}}
\newcommand{\tbblue}[1]{\textbf{\textcolor{color3}{#1}}}

\newcommand{\pub}[1]{{\color{gray}{\tiny{[{#1}]}}}}

\begin{document}
\title{Label-efficient Semantic Scene Completion with Scribble Annotations}


\author{
Song Wang$^1$
\and
Jiawei Yu$^1$\and
Wentong Li$^{1}$\and
Hao Shi$^1$\and
Kailun Yang$^3$\and\\
Junbo Chen$^{2}$\footnotemark[1]\And
Jianke Zhu$^{1}$\footnotemark[1]  %
\affiliations
$^1$Zhejiang University\\
$^2$Udeer.ai\\
$^3$Hunan University\\
\emails 
{\tt\small  jkzhu@zju.edu.cn, junbo@udeer.ai}
}


\twocolumn[{%
\renewcommand\twocolumn[1][]{#1}%
\maketitle
\vspace{-2.3em}
\begin{center}
    \centering
    \includegraphics[width=0.999\textwidth]{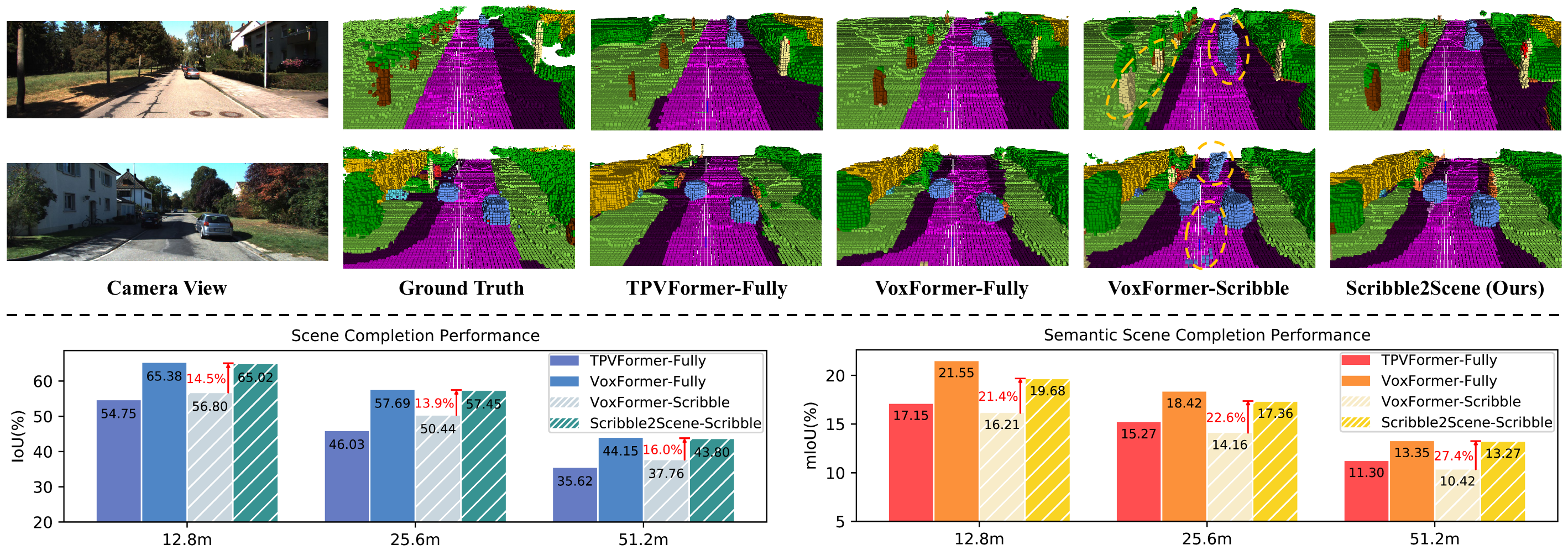}
\vspace{-1.5em}
\captionof{figure}{\textbf{Comparisons between fully supervised methods and our proposed weakly scribble-supervised Scribble2Scene approach} for semantic scene completion on SemanticKITTI. The top half shows examples of semantic occupancy predictions. The bottom half indicates that 
our presented scribble-supervised approach achieves 99\% performance (mIoU) of the fully-supervised methods, which significantly improves the baseline model.}
\label{fig:intro_vis}
\end{center}
}]

\renewcommand{\thefootnote}{\fnsymbol{footnote}}
\footnotetext[1]{Corresponding authors.}
\renewcommand{\thefootnote}{\arabic{footnote}}

\begin{abstract}
  Semantic scene completion aims to infer the 3D geometric structures with semantic classes from camera or LiDAR, which provide essential occupancy information in autonomous driving. Prior endeavors concentrate on constructing the network or benchmark in a fully supervised manner. While the dense occupancy grids need point-wise semantic annotations, which incur expensive and tedious labeling costs. In this paper, we build a new label-efficient benchmark, named ScribbleSC, where the sparse scribble-based semantic labels are combined with dense geometric labels for semantic scene completion. In particular, we propose a simple yet effective approach called Scribble2Scene, which bridges the gap between the sparse scribble annotations and fully-supervision. Our method consists of geometric-aware auto-labelers construction and online model training with an offline-to-online distillation module to enhance the performance. Experiments on SemanticKITTI demonstrate that Scribble2Scene achieves competitive performance against the fully-supervised counterparts, showing 99\% performance of the fully-supervised models with only 13.5\% voxels labeled. Both annotations of ScribbleSC and our full implementation are available at \href{https://github.com/songw-zju/Scribble2Scene}{https://github.com/songw-zju/Scribble2Scene}.
\end{abstract}

\section{Introduction}
Semantic scene completion (SSC), also known as semantic occupancy estimation, aims to predict 3D \textit{geometric} and \textit{semantic} information about the whole scene. Recent studies have shown that occupancy grid can model the environments in a general representation~\cite{tian2023occ3d} and provide essential guidance for downstream tasks~\cite{tong2023scene,liu2024mgmap}, which motivate a series of camera-based methods~\cite{cao2022monoscene,li2023voxformer}.

Current research on semantic scene completion mainly focuses on designing effective network structures under the fully supervised settings~\cite{cao2022monoscene,li2023voxformer,xia2023scpnet}, while few research work provides the solution to learn from less expensive sparse labels.
Meanwhile, existing semantic scene completion~\cite{behley2019semantickitti,li2023sscbench} or occupancy estimation benchmarks~\cite{wang2023openoccupancy,wei2023surroundocc} all heavily rely on fully point-wise annotations for semantic segmentation on LiDAR point cloud, which not only incurs the expensive and tedious manual labeling but also limits their application in new scenarios.
In this work, we firstly revisit the annotation on 3D semantic scene completion, and then propose the scribble-supervised paradigm for this task.

The ground truth for semantic scene completion contains both \textit{geometric} and \textit{semantic} parts that are obtained by accumulating multi-scan point clouds with point-wise semantic labels~\cite{behley2019semantickitti,wang2022meta}. 
The \textit{geometric} part can be directly extracted from raw LiDAR scans while the \textit{semantic} one needs densely annotated labels. We aim to construct a label-efficient benchmark with only sparse semantic labels. Besides, previous efforts~\cite{li2023voxformer,xia2023scpnet,shi2024occfiner} indicate that it is challenging for fully-supervised models to estimate geometry. It is even more severe to learn from sparse semantic labels, as most of the non-empty voxels are unlabeled without semantic information. 
To tackle this issue, we investigate the potential of \textit{dense} geometry offline, thus significantly mitigating the reliance on dense semantic labels. 

In this work, we make full use of the sparse annotations in ScribbleKITTI~\cite{unal2022scribble} to generate scribble-based semantic occupancy labels combined with the dense geometric structure to construct a new benchmark called ScribbleSC.
Specially, we develop a simple yet effective approach, dubbed as Scribble2Scene, which is the first weakly-supervised scheme for 3D semantic occupancy estimation. More importantly, it achieves similar performance compared with the existing fully supervised methods. 
Our proposed Scribble2Scene consists of two stages, including geometry-aware auto-labelers construction (Stage-I) and online model training with distillation (Stage-II).

At Stage-I, we construct geometry-aware auto-labelers with scribble annotations, including Dean-Labeler and Teacher-Labeler. Dean-Labeler treats the complete geometric structure as input, which converts this task into an easier semantic segmentation problem to obtain high-quality voxel-wise segmentation results.
Teacher-Labeler is also trained in offline mode with both input image and complete geometry, which has the same network architecture as the online model. It has the capability to extract more accurate features and semantic logits for the online model. 
At Stage-II, we train the online completion network in a fully-supervised manner based on the pseudo labels provided by Dean-Labeler. In particular, a new range-guided offline-to-online distillation scheme is proposed for large-scale semantic scene understanding, which enhances the performance of the online model with the features from the trained Teacher-Labeler.
Fig.~\ref{fig:intro_vis} shows some qualitative results and comparisons. 

Our main contributions are summarized as below: 
\begin{itemize}
    \item We revisit the annotation of semantic scene completion and propose a scribble-based label-efficient benchmark named ScribbleSC, which provides both sparse semantic annotations and dense geometric labels.
    \item We propose Scribble2Scene, the first weakly-supervised approach for semantic scene completion, designed to handle sparse scribble annotations. Geometry-aware auto-labelers construction and offline-to-online distillation training are devised to accurately predict 3D semantic occupancy.
    \item Under our presented Scribble2Scene framework, the camera-based scribble-supervised model achieves up to a competitive 99\% performance of the fully-supervised one on SemanticKITTI without incurring the computational cost during inference. Additional experiments on SemanticPOSS demonstrate the generalization capability and robustness of our proposed scheme.
\end{itemize}

\section{Related Work}

\noindent \textbf{Semantic Scene Completion.}
Semantic scene completion (SSC) is firstly proposed in SSCNet~\cite{song2017semantic} to construct the complete 3D occupancy with voxel-wise semantic labels from a single-view observation. At the early stage, researchers mainly focus on the indoor scenarios~\cite{liu2018see,zhang2018efficient,li2020attention,cai2021semantic} with RGB image or depth map as input. SemanticKITTI~\cite{behley2019semantickitti} provides the first large dataset and benchmark in the outdoor for autonomous driving.
The subsequent works mainly utilize the occupancy grid voxelized from the current LiDAR frame~\cite{roldao2020lmscnet,wilson2022motionsc} or point cloud directly~\cite{yan2021sparse,cheng2021s3cnet,xia2023scpnet,mei2023ssc} as input and obtain promising performance. Recently, camera-based methods~\cite{cao2022monoscene,huang2023tri,li2023voxformer,zhang2023occformer,yao2023ndc} attract more research attention due to their lower sensor costs. 
VoxFormer~\cite{li2023voxformer} estimates the coarse geometry firstly and adopts the non-empty proposals to perform deformable cross-attention~\cite{zhu2020deformable} with image features, which achieves the best performance among camera-based models.
Along this line, we mainly focus on~\textit{vision only} methods.

\noindent \textbf{Sparsely Annotated Learning.}
Sparse annotations for 2D image segmentation are widely explored including scribble~\cite{lin2016scribblesup,liang2022tree,li2024label}, box~\cite{tian2021boxinst,li2022box,li2024box2mask}, point~\cite{bearman2016s,fan2022pointly,li2023point2mask} and etc.
In 3D scene understanding, ScribbleKITTI~\cite{unal2022scribble} re-annotates the KITTI Odometry dataset~\cite{geiger2012we} and provides the scribble-supervised benchmark for LiDAR segmentation on SemanticKITTI~\cite{behley2019semantickitti}. Box2Mask~\cite{chibane2022box2mask} adopts 3D bounding boxes to train dense segmentation models and achieves 97\% performance of current fully-supervised models. 
We explore the potential of sparse annotations on the geometrically and semantically challenging task of semantic scene completion.

\noindent \textbf{Teacher-Student Network.}
Knowledge Distillation (KD) is proposed initially to transfer the dark knowledge from a large trained teacher model to a small student one for model compression~\cite{hinton2015distilling}.
Following researchers achieve this goal in 2D image at not only logit-level~\cite{cho2019efficacy,furlanello2018born,zhao2022decoupled,liu2022learning} but also the feature-level~\cite{romero2014fitnets,heo2019comprehensive,heo2019knowledge,yang2022masked} for in-depth exploration. 
Teacher-Student architectures in KD are widely adopted in various tasks and applications~\cite{ye2022dynamic,ye2023continual,wang2023lidar2map,wang2024not}.
In SSC, SCPNet~\cite{xia2023scpnet} is proposed to distill dense knowledge from a multi-scan model to a single one with pairwise relational information while its design is exclusively tailored for LiDAR-based methods. CleanerS~\cite{wang2023semantic} generates a perfect visible surface with ground truth voxels and trains a teacher model having cleaner knowledge in indoor scenarios. 
In this paper, we present a new offline-to-online distillation scheme, which is specially designed for 3D semantic scene completion in self-driving environments.

\begin{figure}[t]
\centering
\includegraphics[width=1.0 \linewidth]{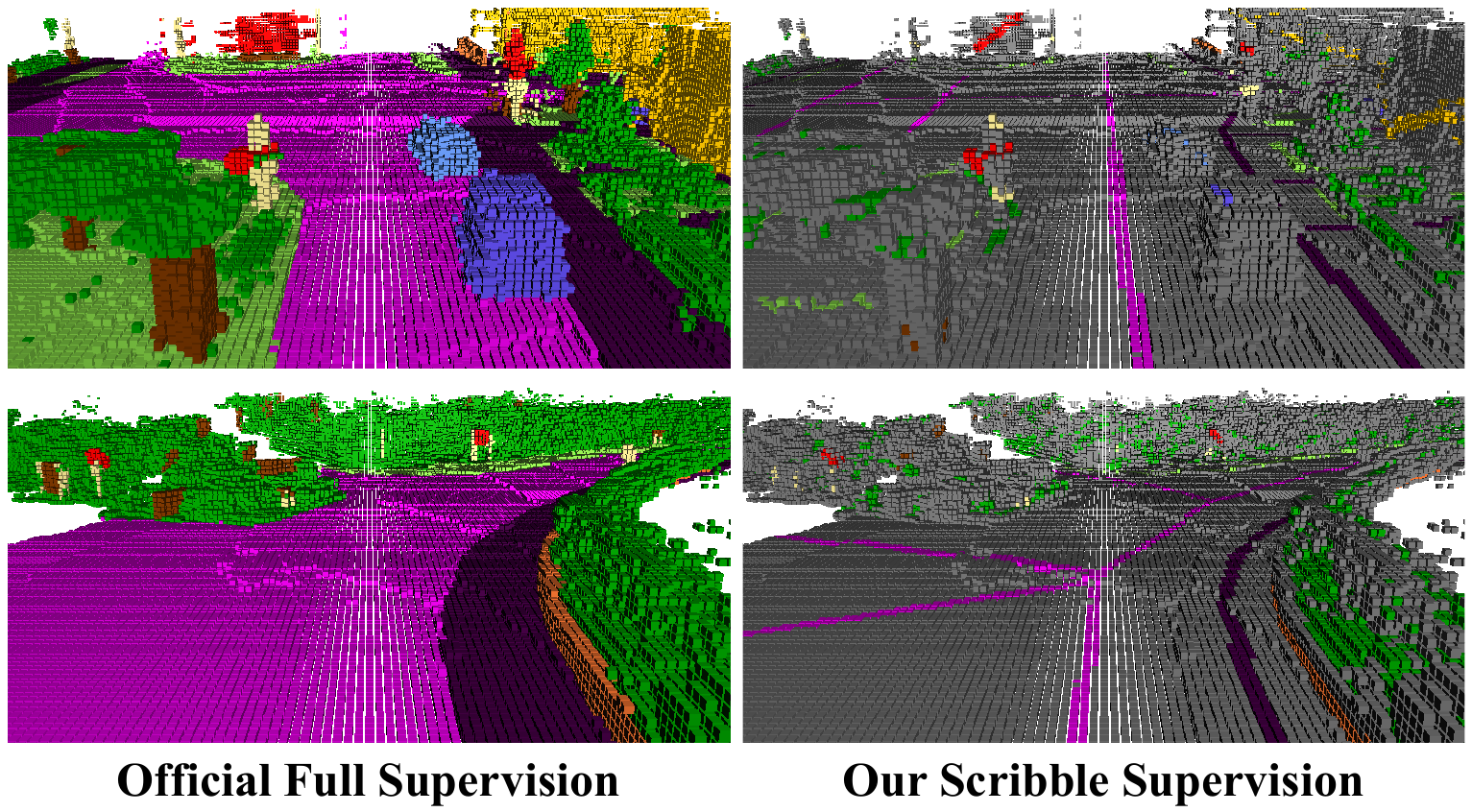}
\vspace{-4mm}
    \caption{Examples of the fully-annotated ground truth from SemanticKITTI (left) and scribble-annotated supervision from our constructed ScribbleSC (right).} 
    \label{fig:example}
    \vspace{-4mm}
\end{figure}

\section{The ScribbleSC Benchmark}
\label{sec:scribble}
The supervision of semantic scene completion can be split into two parts, including \textit{geometric} structure and \textit{semantic} label. 
The \textit{geometric} information can be easily obtained by accumulating exhaustive LiDAR scans and voxelizing the points that fall on the predefined region in front of the car. Meanwhile, the \textit{semantic} label of each voxel is determined by the majority of labeled points within the voxel. 
The annotation of \textit{semantic} part is highly dependent on the dense point-wise semantic segmentation labels, which requires an expensive and complicated labeling process. 
To construct a label-efficient benchmark, we make use of sparse annotations from ScribbleKITTI~\cite{unal2022scribble} to replace the original full annotations provided by SemanticKITTI~\cite{behley2019semantickitti} and achieve the construction of the \textit{semantic} part. In ScribbleKITTI~\cite{unal2022scribble}, line-scribbles are adopted to label the accumulated point clouds and cover only 8.06\% labeled points of the total training set including 10 sequences, which contribute to a 90\% time saving\footnote{Scribble labels cost around 10-25 minutes per tile, while full annotations cost 1.5-4.5 hours.}. 

\noindent \textbf{Label Construction.} Like SemanticKITTI~\cite{behley2019semantickitti}, we superimpose $70$ future LiDAR scans to get the dense geometric structure and choose the volume of $51.2 \,\text{m}$ in the forward of car, $25.6 \,\text{m}$ to the left/right side and $6.4 \,\text{m}$ in height with an off-the-shelf voxelizer tool\footnote{https://github.com/jbehley/voxelizer, MIT License}. The voxel resolution is set to $0.2 \,\text{m}$ and a volume $\mathcal{V}$ of $256 \times 256 \times 32$ voxels can be obtained. We assign the \textit{empty} label to voxels that are devoid of any points. For non-empty voxels, the semantic label is ascertained by conducting a majority vote across the scribble labels of the points situated within the voxel.
If no labeled points exist in a non-empty voxel, we annotate it as \textit{unlabeled}. 
Fig.~\ref{fig:example} provides typical examples from ScribbleSC compared with the fully-annotated SemanticKITTI~\cite{behley2019semantickitti}.
Further information on the label construction for ScribbleSC is included in the Supplementary Material.

\noindent \textbf{Label Usage Instruction.}
The label $\mathcal{V} \in  \mathbb{R}^{X \times Y \times Z}$ ($X, Y, Z = 256, 256, 32$) in ScribbleSC contains 19 semantic classes, an empty class, and an unlabeled class, which can be split into full geometric annotation $\mathcal{G}$ and sparse semantic annotation $\mathcal{S}$. 
The geometric annotation $\mathcal{G} \in  \mathbb{R}^{X \times Y \times Z}$ is a binary voxel grid map. Each voxel is marked as $0$ if it is empty else $1$ for occupied no matter with labeled or unlabeled points. 
The semantic annotation $\mathcal{S} \in  \mathbb{R}^{X \times Y \times Z}$ contains a small number of voxels that are labeled with semantic classes.
The vast majority of voxels in  $\mathcal{S}$ are $0$, including empty and unlabeled ones. 
Training directly with ScribbleSC inevitably encounters an imbalance between geometric and semantic supervision. We will introduce our solution in Sec.~\ref{sec:our_method}.

\begin{figure}[t]
\centering
\includegraphics[width=1.0\linewidth]{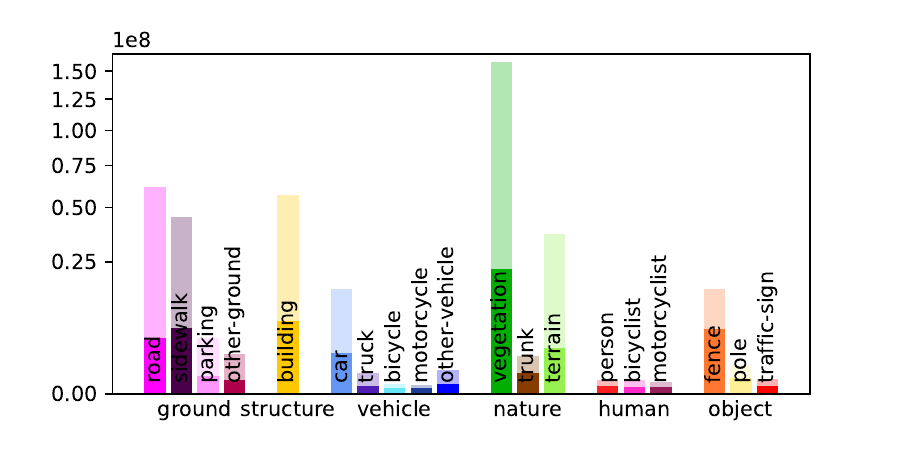}
\vspace{-4mm}
    \caption{Quantity on each category of voxels labeled within ScribbleSC (deep color) in comparison to the fully-annotated SemanticKITTI dataset (light color). The total number of labeled voxels in ScribbleSC is only 13.5\% over SemanticKITTI.}
    \label{fig:stat}
    \vspace{-4mm}
\end{figure}

\noindent \textbf{Voxel Labeling Statistics.}
We have conducted the distribution analysis of semantic labels in ScribbleSC and compared it with the fully annotated SemanticKITTI Benchmark as shown in Fig.~\ref{fig:stat}. Our proposed semantic annotations only contain 13.5\% labeled voxels over SemanticKITTI.
ScribbleSC is a more challenging benchmark as it not only contains a substantial count of empty voxels but also a large number of unlabeled voxels among non-empty ones. More statistical analyses are given in the Supplementary Material.

\section{Proposed Method}
\label{sec:our_method}
\subsection{Overview of Scribble2Scene}

\begin{figure*}[t]
\centering
\includegraphics[width=1.0\linewidth]{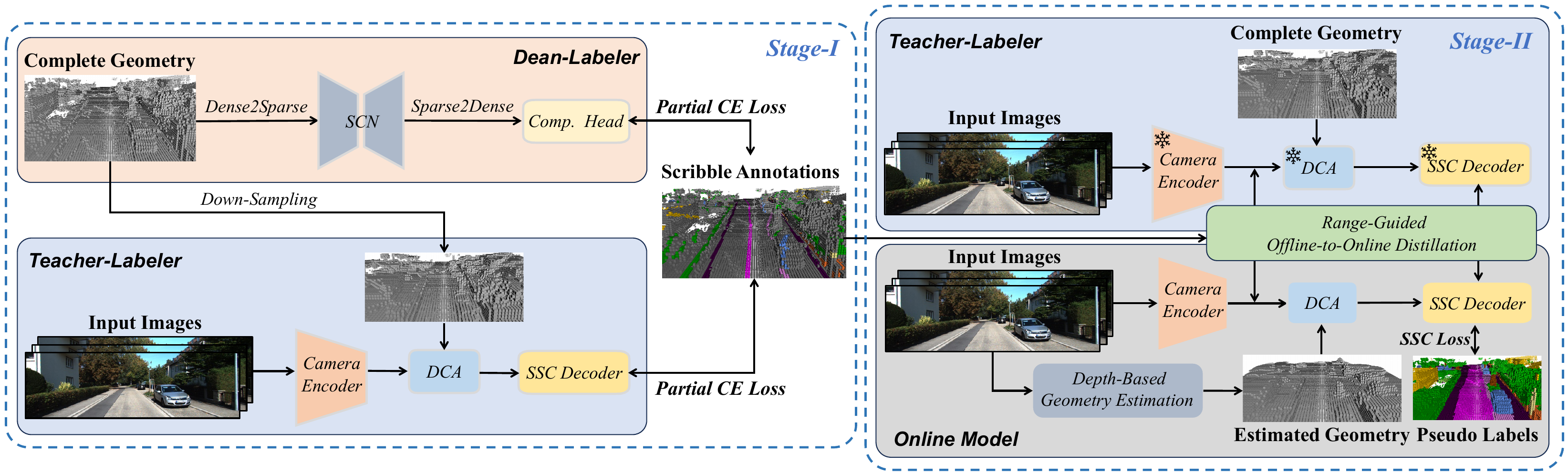}
   \vspace{-6mm}
    \caption{Overview of Scribble2Scene. The left half illustrates the offline \textbf{geometry-aware auto-labelers construction} at Stage-I. The right half shows the \textbf{online model training with distillation} at Stage-II. The accurate pseudo labels from Dean-Labeler and the well-trained Teacher-Labeler are fully leveraged for online model optimization.}
    \label{fig:framework}
    \vspace{-4mm}
\end{figure*}

Fig.~\ref{fig:framework} provides the overview of our proposed Scribble2Scene method for scribble-supervised semantic scene completion, which can be divided into two stages, \textit{i.e.} geometry-aware auto-labelers construction (Stage-I) and online model training with distillation (Stage-II). 
At Stage-I, we make full use of complete geometric structure $\mathcal{G}$ and image $\mathcal{I} \in  \mathbb{R}^{H \times W \times 3}$ observed from the current frame to construct Dean-Labeler and Teacher-Labeler with sparse scribble annotations, as shown in the left of Fig.~\ref{fig:framework}.
At Stage-II, we then adopt the pseudo labels generated by Dean-Labeler and perform the presented range-guided offline-to-online distillation with Teacher-Labeler to train the online model, as illustrated in the right of Fig.~\ref{fig:framework}. 
All the models need to predict the complete 3D occupancy information $\mathcal{O} \in  \mathbb{R}^{X \times Y \times Z \times C}$ in the predefined voxel space,  where $C$ is the number of total categories including empty and semantic classes.
VoxFormer~\cite{li2023voxformer} is employed as the baseline model, which is the state-of-the-art (SOTA) semantic scene completion network with only camera input.
We delve into the specifics of Stage-I in Sec.~\ref{sec:ga2l} and Stage-II in Sec.~\ref{sec:rgo2d}.

\subsection{Geometry-Aware Auto-Labelers}
\label{sec:ga2l}

To fully investigate the potential of the dense geometric structure and sparse semantic labels, we construct two geometry-aware auto-labelers (GA$^2$L) in the offline mode at Stage-I. 
The offline mode means that we can leverage the complete geometry $\mathcal{G}$ from the whole sequences to train a more performant model.

\noindent \textbf{Dean-Labeler.}
Existing SOTA semantic scene completion models~\cite{li2023voxformer,mei2023ssc,xia2023scpnet} employ different branches to process geometric structure and semantic information, respectively. Moreover, their overall performance is often greatly limited by the inaccuracy of geometry estimation.
If we directly treat the complete geometry $\mathcal{G}$ as input, the semantic scene completion can be converted into a voxel-wise semantic segmentation problem. 
Motivated by this, we adopt the sparse convolutional network (\textit{SCN}) as the backbone of Dean-Labeler to obtain voxel-wise semantic prediction from the complete geometry, as illustrated in the top left of Fig.~\ref{fig:framework}. Since the \textit{SCN} does not change the geometric structure of the input, we only need to process the semantic part with scribble supervision.
The partial cross-entropy loss function $\texttt{CE}(\cdot, \cdot)$ to train Dean-Labeler is formulated as 
\begin{equation}
\mathcal{L}_\text{partial\_ce} = \sum_{i=1}^{X} \sum_{j=1}^{Y} \sum_{k=1}^{Z}  \mathcal{S}_{i,j,k} \cdot \texttt{CE}({\mathcal{O}}_{i, j,k}^{(D)}, \mathcal{S}_{i,j,k}) ,
\label{eqn:pce}
\end{equation}
where $\mathcal{O}^{(D)}$ is the predicted occupancy information from Dean-Labeler. We only perform optimization on voxels that are non-empty and labeled. 

Leveraging complete geometry input, Dean-Labeler obtains the promising 3D semantic predictions $\mathcal{P}^{(D)}$(Sec.~\ref{sec:abl}), which eliminates the imbalance between geometry (\textit{dense}) and scribble-based semantic supervision (\textit{sparse}) with ScribbleSC. 
The high-quality semantic predictions serve as the pseudo-labels for online model training at Stage-II. 

\noindent \textbf{Teacher-Labeler.}
To further explore the role of complete geometry $\mathcal{S}$, we design Teacher-Labeler with the modality-specific model VoxFormer~\cite{li2023voxformer}. 
We replace the noisy coarse geometry $\hat{\mathcal{G}_{\frac{1}{2}}} \in \mathbb{R}^{\frac{X}{2} \times \frac{Y}{2} \times \frac{Z}{2}} $ estimated from depth prediction as in VoxFormer with the down-sampling complete geometry ${\mathcal{G}_{\frac{1}{2}}} \in \mathbb{R}^{\frac{X}{2} \times \frac{Y}{2} \times \frac{Z}{2}}$. 
The deformable cross attention (\textit{DCA}) is employed to sample image features with the precise non-empty proposals, as shown in the bottom left of Fig.~\ref{fig:framework}. The \textit{SSC Decoder} is made of a deformable self-attention (\textit{DSA}) module and a completion head. 
The partial cross entropy loss in Eq.~\ref{eqn:pce} is also adopted to train Teacher-Labeler. Additionally, we employ the geometric loss from MonoScene~\cite{cao2022monoscene} with $\mathcal{G}$ to alleviate the geometry change  from the non-empty proposals when 
dense convolutions are used in the completion head.

Teacher-Labeler is trained with the precise non-empty proposals provided by complete geometry, which is able to extract more accurate features from input images. Therefore, the completion model can focus on the semantic part to make this task easier with only sparse scribble semantic annotations.
In our experiments (Sec.~\ref{sec:abl}), Teacher-Labeler achieves extremely higher performance compared to online models using noisy coarse geometry, which is further leveraged for online model training at Stage-II. 

\subsection{Online Model Training with Distillation}
\label{sec:rgo2d}

At Stage-II, we perform the online model training, which only employs currently observed information as inference input. 
The accurate pseudo labels provided by Dean-Labeler are utilized to replace the scribble labels so that we can optimize the model in a fully-supervised paradigm.

The online model as the student network has the same architecture as Teacher-Labeler while it cannot use the complete geometry ${\mathcal{G}}$ as input. To take advantage of the features and predictions obtained from models with the full geometry, we propose a novel range-guided offline-to-online distillation (RGO$^2$D) module that instructs the online model to learn auxiliary modality-specific knowledge at Stage-II. As shown in the right half of Fig.~\ref{fig:framework}, we adopt the well-trained Teacher-Labeler as the offline teacher model and freeze the network weights to perform offline-to-online distillation for the online model, which alleviates the interference of inaccurate pseudo-labels on training. 

\noindent \textbf{Range-Guided Offline-to-Online Distillation.}
When there is a large difference in the input and network performance of the teacher and student models, directly minimizing the Kullback-Leibler (KL) divergence or other metrics to align the outputs, \textit{i.e.} $\mathcal{O}^{(S)} \in  \mathbb{R}^{X \times Y \times Z \times C}$ from the student and $\mathcal{O}^{(T)} \in  \mathbb{R}^{X \times Y \times Z \times C}$ from the teacher, often does not work well~\cite{huang2022knowledge}. Inspired by CleanerS~\cite{wang2023semantic}, we adopt global semantic logit combined with local semantic affinity rather than the original predictions to perform distillation between the teacher and student at the logit level. Considering that the outdoor driving scene involves a wider range, 
only using the semantic logit and affinity of the whole scene cannot reflect the distribution of each class well. Moreover, the semantic logit and affinity of the area closer to the ego-vehicle are inherently more amenable to learning processes and endow more significance in ensuring vehicular safety assurances.
Therefore, we consider introducing range information as guidance to model the distribution of each semantic class.

\begin{figure}[t]
\centering
\includegraphics[width=1.0\linewidth]{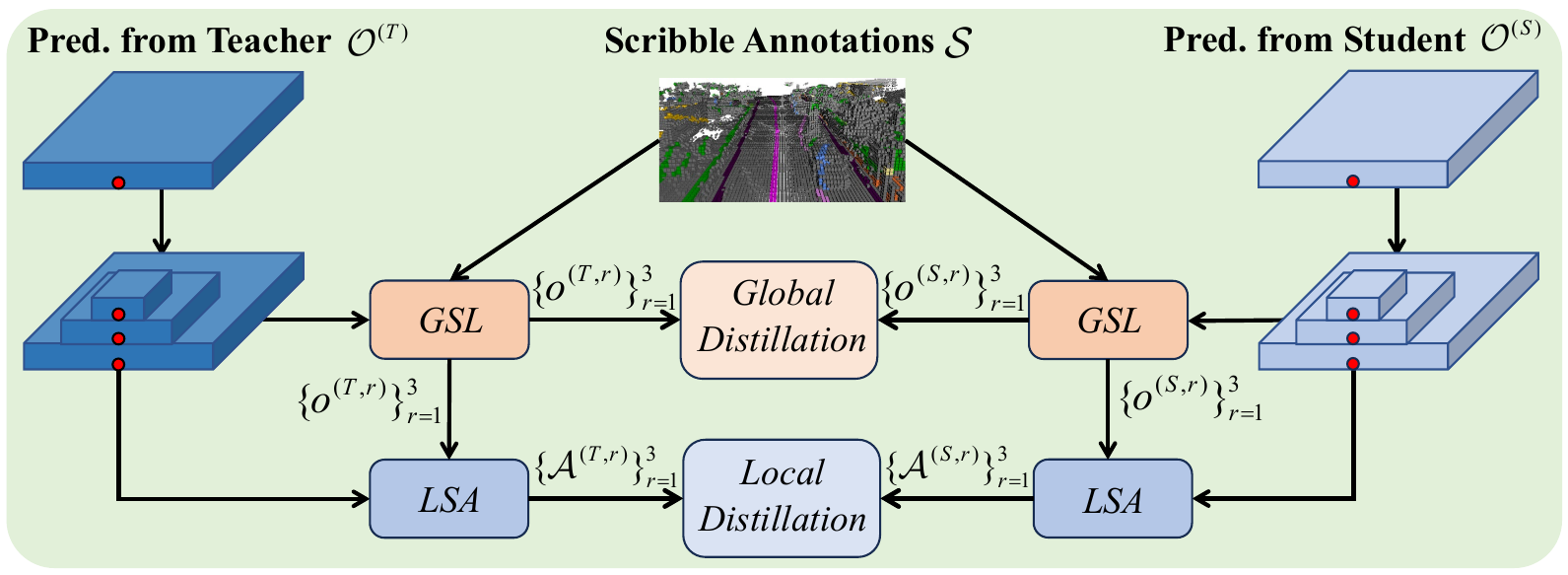}
    \vspace{-6mm}
    \caption{Illustration of \textbf{range-guided offline-to-online distillation} scheme. The red dot denotes the location of the ego-vehicle. The global and local distillation with different ranges are performed, respectively.} 
    \label{fig:rgdistill}
    \vspace{-4mm}
\end{figure}

Specifically, we take the ego-vehicle as the center and divide the whole volume in the predictions of teacher $\mathcal{O}^{(T)}$ and student $\mathcal{O}^{(S)}$ into three ranges (including near, middle and far) according to the distance as shown in Fig.~\ref{fig:rgdistill}. Then the global semantic logits (\textit{GSL}) of teacher $\{{\boldsymbol{o}}^{(T,r)}\}_{r=1}^{3}$ and student $\{{\boldsymbol{o}}^{(S,r)}\}_{r=1}^{3}$ at different ranges are calculated with scribble annotations $\mathcal{S}$. 
To adequately measure the difference of global semantic logits between student and teacher, we introduce the inter-relation and intra-relation loss to perform global distillation
\begin{small}
\begin{equation}
\mathcal{L}_{\text {global}, r} \! =  \!
 \frac{1}{C}  [ \alpha \! \sum_{i=1}^C \! d_{\mathrm{p}}  (\boldsymbol{o}_{i,:}^{(S, r)}\!,\! \boldsymbol{o}_{i,:}^{(T,r)} ) 
 \! + \! \beta  \! \sum_{j=1}^C \! d_{\mathrm{p}}  (\boldsymbol{o}_{:, j}^{(S, r)}\!,\! \boldsymbol{o}_{:, j}^{(T, r)}) ] ,
\end{equation}
\end{small}
\noindent where $d_{\mathrm{p}}$ is Pearson's distance, and $r=1,2,3$ denote different ranges. $\alpha$ and $\beta$ are the balanced weights of the inter- and intra-relation loss, respectively.
Moreover, we compute the local semantic affinity (\textit{LSA}) with the global semantic logit and the prediction at each range for teacher $\{\mathcal{A}^{(T, r)}\}_{r=1}^{3}$ and student $\{\mathcal{A}^{(S, r)}\}_{r=1}^{3}$. Then local distillation loss is computed below
\begin{equation}
\mathcal{L}_{\text{local}, r}= \texttt{MSE}(\mathcal{A}^{(S, r)},\mathcal{A}^{(T, r)}) ,
\end{equation}
where \texttt{MSE}$(\cdot,\cdot)$ denotes the mean square error function.

Additionally, we adopt \texttt{MSE}$(\cdot,\cdot)$ as the feature-level distillation loss $\mathcal{L}_{\text{feat}}$. We choose the 2D features from the image encoder as the targets to align. 
Overall, the range-guided offline-to-online distillation is composed of the above items
\begin{equation}
\mathcal{L}_{\text{distill}}=\mathcal{L}_{\text{feat}} + \sum_{r=1}^{3} w_r \cdot (\mathcal{L}_{\text{global,r}} + \mathcal{L}_{\text{local},r}) ,
\end{equation}
where $w_r$ denotes the range loss coefficients, and ${r}=1, 2, 3$. $\mathcal{L}_{\text{global,r}}$ and $\mathcal{L}_{\text{local,r}}$ represent the global and local distillation losses at near, middle, and far range, respectively.

\begin{table*}[!htb]\centering
\renewcommand\tabcolsep{5pt}
\scriptsize
\begin{tabular}{l|ccc|ccc|ccc|ccc|ccc}\toprule
\textbf{Methods} &\multicolumn{3}{c|}{\textbf{Scribble2Scene (Ours)}} &\multicolumn{3}{c|}{\textbf{VoxFormer$^{\dag}$}~\pub{CVPR'23}} &\multicolumn{3}{c|}
{\textbf{VoxFormer}~\pub{CVPR'23}} &\multicolumn{3}{c|}{\textbf{TPVFormer}~\pub{CVPR'23}} &\multicolumn{3}{c}{\textbf{MonoScene$^{*}$}~\pub{CVPR'22}} \\\midrule
\textbf{Supervision} &\multicolumn{3}{c|}{\textbf{Scribble}} &\multicolumn{3}{c|}{\textbf{Scribble}} &\multicolumn{3}{c|}{\textbf{Fully}} &\multicolumn{3}{c|}{\textbf{Fully}} &\multicolumn{3}{c}{\textbf{Fully}} \\\midrule
\textbf{Range} &\textbf{12.8m} &\textbf{25.6m} &\textbf{51.2m} &\textbf{12.8m} &\textbf{25.6m} &\textbf{51.2m}  &\textbf{12.8m} &\textbf{25.6m} &\textbf{51.2m}  &\textbf{12.8m} &\textbf{25.6m} &\textbf{51.2m}  &\textbf{12.8m} &\textbf{25.6m} &\textbf{51.2m}  
\\\midrule
\crule[carcolor]{0.13cm}{0.13cm} \textbf{car} (3.92\%) &\tbblue{{40.06}} &\tbblue{{33.15}} &\tbblue{{24.16}} & 31.02 & 24.99 & 17.77& \tbb{44.90} & \tbb{37.46} &\tbb{26.54}  & 34.81 & 31.72 & 23.79 &24.34 &24.64 &{23.29}  \\
\crule[bicyclecolor]{0.13cm}{0.13cm} \textbf{bicycle} (0.03\%) &{0.12} &{0.57} &{0.30}& \tbblue{0.78} & \tbblue{0.59 } & \tbblue{0.33} &\tbb{5.22} & \tbb{2.87} & \tbb{1.28}  & 0.33 & 0.69 & 0.35 &{0.07} &{0.23} &{0.28}   \\
\crule[motorcyclecolor]{0.13cm}{0.13cm} \textbf{motorcycle} (0.03\%)&\tbblue{{3.62}} &\tbblue{{1.45}} &\tbblue{{1.01}} & 0.05 & 0.02 & 0.03 & \tbb{2.98} & \tbb{1.24} &{0.56}  & 0.16 & 0.08 & 0.05 &{0.05} &{0.20} & \tbb{0.59}   \\
\crule[truckcolor]{0.13cm}{0.13cm} \textbf{truck} (0.16\%) &\tbblue{{14.32}} &\tbblue{{17.06}} &\tbblue{{17.32}} & 7.39 & 5.18 & 4.35 & {9.80} &{10.38} &{7.26}  & \tbb{17.77} & 13.15 & 6.92 &{15.44} & \tbb{13.84} & \tbb{9.29}  \\
\crule[othervehiclecolor]{0.13cm}{0.13cm} \textbf{other-veh.} (0.20\%) &\tbblue{{11.59}} &\tbblue{{5.76}} &\tbblue{{3.69}} & 4.02 &  1.51 & 0.87 & \tbb{17.21} & \tbb{10.61} & \tbb{7.81}  & 10.06 & 7.47 & 4.29 &{1.18} &{2.13} &{2.63}   \\
\crule[personcolor]{0.13cm}{0.13cm} \textbf{person} (0.07\%) &\tbblue{{5.01}} &{3.53} &{1.98} & 4.02 & \tbblue{3.81} & \tbblue{2.27} & \tbb{4.44} & \tbb{3.50} &{1.93}  &  1.56 & 1.06 & 0.52 &{0.90} &{1.37} & \tbb{2.00}  \\
\crule[bicyclistcolor]{0.13cm}{0.13cm} \textbf{bicyclist} (0.07\%)&{1.33} &{1.08} &{0.47} & \tbblue{3.59} & \tbblue{4.85} & \tbblue{2.67} & \tbb{2.65} & \tbb{3.92} & \tbb{1.97}  & 2.57 & 1.93 &  0.91 &{0.54} &{1.00} &{1.07}   \\
\crule[motorcyclistcolor]{0.13cm}{0.13cm} \textbf{motorcyclist} (0.05\%) & 0.00 & 0.00 & 0.00 & 0.00 & 0.00 & 0.00 &0.00 &0.00 &0.00  &  0.00 & 0.00 & 0.00 &0.00 &0.00 &0.00   \\
\crule[roadcolor]{0.13cm}{0.13cm} \textbf{road} (15.30\%) &\tbblue{{68.05}} &\tbblue{{60.75}} &\tbblue{{49.90}} & 61.11 & 54.56 & 45.23 &{75.45} &{66.15} &{53.57}  & \tbb{75.91} &  \tbb{69.42} & \tbb{56.47} &57.37 &57.11 &{55.89}   \\
\crule[parkingcolor]{0.13cm}{0.13cm} \textbf{parking} (1.12\%) &\tbblue{{20.88}} &\tbblue{{22.47}} &\tbblue{{20.12}} & 14.45 & 16.52 & 16.19 &{21.01} &{23.96} &{19.69}  & \tbb{29.88} & \tbb{26.22} & \tbb{20.59} &{20.04} &{18.60} &{14.75}  \\
\crule[sidewalkcolor]{0.13cm}{0.13cm} \textbf{sidewalk} (11.13\%) &\tbblue{{44.43}} &\tbblue{{35.71}} &\tbblue{{26.93}} & 37.88 & 30.27 & 20.67 &{45.39} &{34.53} & \tbb{26.52}  & \tbb{47.05} & \tbb{36.69} & 25.83 &27.81 &27.58 &{26.50}   \\
\crule[othergroundcolor]{0.13cm}{0.13cm} \textbf{other-grnd}(0.56\%) & 0.00 &\tbblue{{0.40}} &\tbblue{{0.87}} & 0.00 & 0.12 & 0.57 & 0.00 &{0.76} &{0.42}  & 0.00 & 1.36 & 0.94 & \tbb{1.73} & \tbb{2.00} & \tbb{1.63}   \\
\crule[buildingcolor]{0.13cm}{0.13cm} \textbf{building} (14.10\%) &\tbblue{{25.76}} &\tbblue{{30.62}} &\tbblue{{20.14}} & 21.41& 25.86 & 16.52 & \tbb{25.13} & \tbb{29.45} & \tbb{19.54}  & 11.37 & 18.23 & 13.89 & {16.67} &15.97 &13.55   \\
\crule[fencecolor]{0.13cm}{0.13cm} \textbf{fence} (3.90\%) &\tbblue{{12.37}} &\tbblue{{8.68}} &\tbblue{{6.12}} & 10.90 & 6.59 &  4.11 & \tbb{16.17} &\tbb{11.15} & \tbb{7.31}  & 9.81 & 7.98 & 5.99 &{7.57} &{7.37} &{6.60}   \\
\crule[vegetationcolor]{0.13cm}{0.13cm} \textbf{vegetation} (39.3\%) &\tbblue{{44.00}} &\tbblue{{38.35}} &\tbblue{{25.99}} & 39.30 & 33.87 & 21.56 & \tbb{43.55} & \tbb{38.07} & \tbb{26.10}  & 24.90 & 24.32 & 16.93 &19.52 &19.68 &17.98   \\
\crule[trunkcolor]{0.13cm}{0.13cm} \textbf{trunk} (0.51\%) &\tbblue{{21.23}}  &\tbblue{{14.00}} &\tbblue{{8.03}} & 17.10 & 11.35& 5.86& \tbb{21.39}  &\tbb{12.75} & \tbb{6.10}  & 8.91 & 4.53 & 2.25 &2.02 &2.57 & 2.44   \\
\crule[terraincolor]{0.13cm}{0.13cm} \textbf{terrain} (9.17\%) &\tbblue{{41.83}} &\tbblue{{38.86}} &\tbblue{{32.39}} & 38.89 & 35.96 & 30.36 & \tbb{42.82} & \tbb{39.61} & \tbb{33.06}  & 41.12 & 38.02 & 30.35 &31.72 &31.59 &29.84  \\
\crule[polecolor]{0.13cm}{0.13cm} \textbf{pole} (0.29\%) &\tbblue{{11.84}}&\tbblue{{10.43}} &\tbblue{{7.52}} &  9.98 & 7.41 & 4.55 & \tbb{20.66} &\tbb{15.56} & \tbb{9.15}  &  7.30 & 4.99  & 3.13 & 3.10 &3.79 &3.91   \\
\crule[trafficsigncolor]{0.13cm}{0.13cm} \textbf{traf.-sign} (0.08\%) &\tbblue{{7.39}} &\tbblue{{6.94}} &\tbblue{{5.25}} & 6.04 & 5.55 & 4.06 & \tbb{10.63} &\tbb{8.09} & \tbb{4.94}  & 2.35 & 2.31 & 1.52 &3.69 & 2.54 &{2.43}   \\\midrule
\textbf{IoU (\%)} & \tbblue{{{65.02}}} &\tbblue{{{57.45}}} &\tbblue{{{43.80}}} & 56.80 & 50.44 & 37.76 & \tbb{65.38} &\tbb{{57.69}} &\tbb{{44.15}}  & 54.75 & 46.03 & 35.62 &38.42 &38.55 &36.80    \\
\textbf{mIoU (\%)} &\tbblue{{19.68}} &\tbblue{{17.36}} &\tbblue{{13.27}} & 16.21 & 14.16 & 10.42 & \tbb{21.55} & \tbb{18.42} & \tbb{13.35}  & 17.15 & 15.27 & 11.30 &12.25 &12.22 &{{11.30}} \\  
\textbf{SS/FS (\%)} & \tbblue{91.32} & \tbblue{94.25} & \tbblue{99.40} & 75.22 & 76.87 & 78.05 &{-} &{-} &-  & - & - & - &- &- &{-} 
\\\bottomrule
\end{tabular}
\vspace{-2.5mm}
\caption{Quantitative comparisons against the fully-supervised camera-based SSC methods on the \textit{\textbf{validation set}} of SemanticKITTI. $\dag$ represents the results that are directly retrained with ScribbleSC. $*$ denotes the results are reported from VoxFormer. SS/FS measures the relative performance ratio of the scribble-supervised (SS) model over the fully-supervised (FS) one. The best results in scribble- and fully-supervised models are marked in \tbblue{blue} and \tbb{bold}, respectively.}
\label{tab:comp_val}
\vspace{-3mm}
\end{table*}

\begin{table*}[htb]
\renewcommand\tabcolsep{2.5pt}
\scriptsize
    \centering
    \begin{tabular}{r|c|c|c|c|c|c|c|c|c|c|c|c|c|c|c|c|c|c|c|c|c|cc}
   \toprule 
\multicolumn{1}{c|}{\textbf{Methods}} &\rotatebox{90}{\textbf{Supervision}}&\rotatebox{90}{\textbf{IoU (\%)}}
& \rotatebox{90}{\crule[carcolor]{0.13cm}{0.13cm} \textbf{car}}
& \rotatebox{90}{\crule[bicyclecolor]{0.13cm}{0.13cm} \textbf{bicycle}}
& \rotatebox{90}{\crule[motorcyclecolor]{0.13cm}{0.13cm} \textbf{motorcycle}} 
& \rotatebox{90}{\crule[truckcolor]{0.13cm}{0.13cm} \textbf{truck}} 
& \rotatebox{90}{\crule[othervehiclecolor]{0.13cm}{0.13cm} \textbf{other-veh.}}
& \rotatebox{90}{\crule[personcolor]{0.13cm}{0.13cm} \textbf{person}}
& \rotatebox{90}{\crule[bicyclistcolor]{0.13cm}{0.13cm} \textbf{bicyclist}}
& \rotatebox{90}{\crule[motorcyclistcolor]{0.13cm}{0.13cm} \textbf{motorcyclist}}
& \rotatebox{90}{\crule[roadcolor]{0.13cm}{0.13cm} \textbf{road}}
& \rotatebox{90}{\crule[parkingcolor]{0.13cm}{0.13cm} \textbf{parking}} 
& \rotatebox{90}{\crule[sidewalkcolor]{0.13cm}{0.13cm} \textbf{sidewalk}}
& \rotatebox{90}{\crule[othergroundcolor]{0.13cm}{0.13cm} \textbf{other-grnd}}
& \rotatebox{90}{\crule[buildingcolor]{0.13cm}{0.13cm} \textbf{building}}
& \rotatebox{90}{\crule[fencecolor]{0.13cm}{0.13cm} \textbf{fence}}
& \rotatebox{90}{\crule[vegetationcolor]{0.13cm}{0.13cm} \textbf{vegetation}}
& \rotatebox{90}{\crule[trunkcolor]{0.13cm}{0.13cm} \textbf{trunk}}
&\rotatebox{90}{\crule[terraincolor]{0.13cm}{0.13cm} \textbf{terrain}}
&\rotatebox{90}{\crule[polecolor]{0.13cm}{0.13cm} \textbf{pole}} 
&\rotatebox{90}{\crule[trafficsigncolor]{0.13cm}{0.13cm} \textbf{traf.-sign}}
&\rotatebox{90}{\textbf{mIoU (\%)}}
\\ \midrule

\textbf{MonoScene}~\pub{CVPR'22} & Fully & 34.16 & 18.80 & 0.50 & 0.70 & {3.30} & {{4.40}} & {1.00} & 1.40 & {{0.40}}& {54.70}& {24.80}& {27.10}& 5.70& 14.40& {11.10}& 14.90&2.40& 19.50& 3.30& 2.10& 11.08 \\
\textbf{TPVFormer}~\pub{CVPR'23} & Fully & {34.25} & 19.20 & 1.00 & 0.50 & {3.70} & 2.30 & {1.10} & \tbb{{2.40}} & 0.30 & 55.10 & 27.40 & 27.20 &{6.50} & 14.80 & 11.00 & 13.90 & 2.60 & 20.40& 2.90 & 1.50 & 11.26\\
\textbf{OccFormer}~\pub{ICCV'23} & Fully & {34.53} & 21.60 & 1.50 & 1.70 & 1.20 & 3.20 & {{2.20}} & {1.10} & 0.20 & {55.90} & \tbb{31.50} & \tbb{30.30} &{6.50} & 15.70 & 11.90 &16.80 & 3.90& 21.30& 3.80 & 3.70 & 12.32\\
\textbf{NDC-Scene}~\pub{ICCV'23} & Fully & {36.19} & 19.13 & \textbf{1.93} & \textbf{2.07} & \textbf{4.77} & \textbf{6.69} & \textbf{{3.44}} & {2.77} & \textbf{1.64} & \textbf{58.12} & {25.31} & {28.05} &{6.53} & 14.90 & 12.85 &17.94 & 3.49 & {25.01} & 4.43 & 2.96 & 12.58\\
\textbf{VoxFormer}~\pub{CVPR'23} & Fully & \tbb{43.21} & \tbb{21.70} & {{1.90}} & {{1.60}} & {3.60} & 4.10& {1.60} & {1.10}& 0.00& 54.10 & {25.10} & 26.90& \tbb{{7.30}} & \tbb{{23.50}}& \tbb{{13.10}}& \tbb{{24.40}}& \tbb{{8.10}}& \tbb{{24.20}}& \tbb{{6.60}}& \tbb{{5.70}}& \tbb{13.41}\\\midrule
\textbf{VoxFormer$^{\dag}$}~\pub{CVPR'23} & Scribble & {34.43} & 17.50 & 1.90 & 1.00 & 1.20 & 4.40 & 1.00 & 1.40 & 0.00 & 44.30 & 18.90 & 23.10 & 10.10 & 17.00 & 8.00 & 18.20 & 7.10 & 23.10 & 3.40 & 4.80 & {10.87}
\\

\textbf{Scribble2Scene (Ours)} & Scribble &  \tbblue{42.60} & \tbblue{20.10} & \tbblue{2.70} & \tbblue{1.60 }& \tbblue{5.60} & \tbblue{4.50} & \tbblue{1.60} & \tbblue{1.80 }& {0.00} & \tbblue{50.30} & \tbblue{20.60} & \tbblue{27.30} & \tbblue{11.30} & \tbblue{23.70} &\tbblue{ 13.30} & \tbblue{23.50} & \tbblue{9.60} & \tbblue{23.80} & \tbblue{5.60} & \tbblue{6.50} & \tbblue{13.33}  \\

\bottomrule
\end{tabular}
\vspace{-2.5mm}
    \caption{Quantitative comparisons between Scribble2Scene and the state-of-the-art camera-based methods on the \textit{\textbf{hidden test set}} of SemanticKITTI. 
    $\dag$ represents the results that are retrained with ScribbleSC.}
    \label{tab:hiddentest}
\vspace{-4mm}
\end{table*}

\subsection{Training and Inference}
\noindent \textbf{Overall Loss.}
The total loss $\mathcal{L}_{\text{total}}$ for the online model training consists of semantic loss $\mathcal{L}_{\text{sem}}$, geometric loss $\mathcal{L}_{\text{geo}}$ and distillation loss $\mathcal{L}_{\text{distill}}$ as below
\begin{equation}
\mathcal{L}_{\text{total}}=\mathcal{L}_{\text{sem}} + \mathcal{L}_{\text{geo}}+\mathcal{L}_{\text{distill}}, 
\end{equation}
where $\mathcal{L}_{\text{sem}}$ is the commonly used weighted cross-entropy loss. We employ the pseudo labels $\mathcal{P}^{(D)}$ from Dean-Labelers as the full semantic supervision.  $\mathcal{L}_{\text{geo}}$ is the geometric scene-class affinity loss proposed in MonoScene~\cite{cao2022monoscene}.

\noindent \textbf{Inference.}
At the inference stage, we only need to preserve the student branch, which can obtain similar performance on accuracy while retaining efficient inference as the fully-supervised model.

\section{Experiments}
\subsection{Experimental Setup}
\noindent \textbf{Dataset.} Our models are trained on ScribbleSC. Unless specified, the performance is mainly evaluated on the \textit{validation set} of the fully-annotated SemanticKITTI~\cite{behley2019semantickitti}, which is a highly challenging benchmark. All input images come from the KITTI Odometry Benchmark~\cite{geiger2012we} consisting of 22 sequences. Following the official setting, we use the sequences 00-10 except 08 for training with ScribbleSC while sequence 08 is preserved as the \textit{validation set}. We submit the predictions of sequences 11-21 to the online evaluation website and obtain the scores on the \textit{hidden test set}. 
Additionally, we have conducted extra experiments on the SemanticPOSS~\cite{pan2020semanticposs}, which is another challenging dataset collected in a campus-based environment. 
Since the scribble-based annotations on point clouds of SemanticPOSS are unavailable, we randomly sample 10\% of its full annotations to obtain similar sparse labels as scribbles. Then we construct semantic scene completion labels, including sparse semantic labels and dense geometric labels as described in Sec.~\ref{sec:scribble}. 
Adhering to the original configuration, the sequences (00-01, 03-05) / 02 are split as \textit{training} and \textit{validation set}, respectively.

\noindent \textbf{Evaluation Protocol.} 
We follow the official evaluation benchmark and employ intersection over union ({IoU}) to evaluate the scene completion performance, which only measures the class-agnostic geometric quality. 
The standard mean intersection over union metric ({mIoU}) of 19 semantic classes is reported for semantic scene completion. We choose the class-wise mIoU as the dominant evaluation metric. 
To comprehensively compare with the fully-supervised methods, we provide evaluation scores from three different ranges on \textit{validation set} including $12.8 \times 12.8 \times 6.4\text{m}^{3}$, $25.6 \times 25.6 \times 6.4\text{m}^{3}$, and $51.2 \times 51.2 \times 6.4\text{m}^{3}$.

\noindent \textbf{Implementation Details.} 
For Dean-Labeler, we adopt Cylinder3D~\cite{zhu2021cylindrical} as the \textit{SCN} backbone and use a single GPU to train the network with a batch size of $4$.
For Teacher-Labeler and student model, we use the same backbone of VoxFormer-T~\cite{li2023voxformer}, which takes the current and previous $4$ images as input. 
All models based on VoxFormer are trained on $4$ GPUs with $20$ epochs, a batch size of $1$ (containing $5$ images) per GPU. 
Our baseline model is directly trained with partial cross-entropy loss and geometric loss under the available scribble- and geometry-based supervisions.
For our proposed range-guided distillation scheme, we choose three different ranges, which are the same as the evaluation part to perform the global and local distillation, \textit{i.e.} $12.8\text{m}$, $25.6\text{m}$, and $51.2\text{m}$.
More implementation details and model complexity analyses are provided in the Supplementary Material.

\subsection{Main Results}
We firstly compare Scribble2Scene with state-of-the-art fully-supervised camera-based methods on the \textit{validation set} of SemanticKITTI, including VoxFormer~\cite{li2023voxformer}, TPVFormer~\cite{huang2023tri}, and MonoScene~\cite{cao2022monoscene}. As shown in Tab.~\ref{tab:comp_val}, Scribble2Scene obtains 99\% performance (13.27\% mIoU \textit{v.s.} 13.35\% mIoU) of fully-supervised VoxFormer at full-range $51.2\text{m}$, which only uses 13.5\% of the labeled voxels.
The competitive accuracy is also achieved against other camera-based models at different ranges. Compared with VoxFormer trained directly using scribble annotations, our method has a significant improvement in the most of categories.
To further examine the effectiveness of our method, we submit results on the extremely challenging \textit{test set} of SemanticKITTI without extra tricks. As illustrated in Tab.~\ref{tab:hiddentest}, our scribble-based method achieves 13.33\% mIoU and 42.60\% IoU, which outperforms most of the fully-supervised models and demonstrates generalization capability in more scenarios.

\begin{table}[t]\centering
\scriptsize
\renewcommand\tabcolsep{3.pt}
{
\begin{tabular}{r|c|cc|c}\toprule
\multicolumn{1}{c|}{\textbf{Methods}} & \textbf{Supervision} & \textbf{IoU (\%)} & \textbf{mIoU (\%)} & \textbf{SS/FS (\%)} \\\midrule
\textbf{LMSCNet}~\pub{3DV'20} & Fully & 54.27 & 16.24 & - \\
\textbf{LMSCNet$^{\dag}$}~\pub{3DV'20} & Sparse & 31.00 & 12.26 & 75.49  \\
\textbf{LMSCNet with S2S (Ours)} & Sparse & 53.24 & \textbf{16.01} & \textbf{98.58} 
\\\midrule
\textbf{MotionSC}~\pub{RA-L'22} & Fully & 53.28 & 18.10 & - \\
\textbf{MotionSC$^{\dag}$}~\pub{RA-L'22} & Sparse & 33.45 & 12.91 &  71.33 \\
\textbf{MotionSC with S2S (Ours)} & Sparse & 53.48 & \textbf{17.63} &  \textbf{97.40} \\
\bottomrule
\end{tabular}
}
\vspace{-2.5mm}
\caption{Quantitative comparisons against the fully-supervised SSC methods on the \textit{validation set} of SemanticPOSS. $\dag$ indicates the results that are directly retrained with the sparse semantic labels.
We also re-implement these methods with our Scribble2Scene (S2S).}
\vspace{-4mm}
\label{tab:sparse_poss}
\end{table}

\begin{table}[t]\centering
\scriptsize
\setlength{\tabcolsep}{2mm}{
\begin{tabular}{r|c|c|cc}\toprule
\multicolumn{1}{c|}{\textbf{Methods}} & \textbf{Input} & \textbf{Supervision} & \textbf{IoU (\%)} & \textbf{mIoU (\%)} \\\midrule
\textbf{SCPNet}~\pub{CVPR'23} & L & Fully & 49.90 & \textbf{37.20} \\
\textbf{S3CNet}~\pub{CoRL'20} & L & Fully & 57.12 & 33.08 \\
\textbf{SSC-RS}~\pub{IROS'23} & L & Fully & {58.62} & 24.75 \\
\textbf{JS3C-Net}~\pub{AAAI'21} & L & Fully & 53.09 & 22.67 \\
\midrule
\textbf{VoxFormer}~\pub{CVPR'23} & C & Fully & {44.15} & \textbf{13.35} \\
\textbf{TPVFormer}~\pub{CVPR'23} & C & Fully & 35.62 & 11.30 \\
\textbf{MonoScene}~\pub{CVPR'22} & C & Fully & 36.80 & 11.30 \\
\midrule
\textbf{Dean-Labeler} & G & Scribble & {100.00} & \textbf{42.28} \\
\textbf{Teacher-Labeler} & C \& G & Scribble & 82.80 & 21.70 \\
\textbf{Scribble2Scene (Ours)} &  C & Scribble & {43.80} & {13.27} \\
\bottomrule
\end{tabular}}
\vspace{-2.5mm}
\caption{Performance of geometric-aware auto-labelers (Stage-I) against the state-of-the-art semantic scene completion models. ``L'', ``C'', and ``G'' denote the LiDAR, camera, and complete geometry, respectively.}
\label{tab:scribble_ann}
\vspace{-4mm}
\end{table}

Additional experiments are conducted on SemanticPOSS with our Scribble2Scene framework.
Since SemanticPOSS only provides point cloud as input, LiDAR-based methods including LMSCNet~\cite{roldao2020lmscnet} and MotionSC~\cite{wilson2022motionsc} are adopted as baseline models. 
As shown in Tab.~\ref{tab:sparse_poss}, our method outperforms the baseline models significantly (16.01\% mIoU \textit{v.s.} 12.26\% mIoU, 17.63\% mIoU \textit{v.s.} 12.91\% mIoU), which showcases its adaptability and robustness across diverse datasets and models.
The implementation details and results on each semantic category are provided in the Supplementary Material.

\begin{table}[t]\centering
\scriptsize
\setlength{\tabcolsep}{3.2mm}{
\begin{tabular}{clccccc}\toprule
\multirow{2}{*}{\textbf{Baseline}} & \multicolumn{3}{c}{\textbf{Scribble2Scene}} &\multirow{2}{*}{\textbf{IoU (\%)}} &\multirow{2}{*}{\textbf{mIoU (\%)}}  \\
\cmidrule(r){2-4}
& DL & TL & RGO$^2$D & & \\ \midrule
\cmark &  &  & & 37.76  & 10.42   \\
\cmark &   & \cmark & & 36.51  & 10.37   \\
\cmark &   &\cmark &\cmark & 36.86 &  10.79 \\
& \cmark & &  & 44.19  & 10.56   \\
& \cmark  &\cmark & & {44.51} & 11.27  \\
& \cmark  &\cmark &\cmark & {43.80} &\textbf{13.27}  \\
\bottomrule
\end{tabular}
}
\vspace{-2.5mm}
\caption{{Impact of the each module in overall Scribble2Scene framework.} }
\label{tab:abl}
\vspace{-4mm}
\end{table}

\begin{table}[t]\centering
\scriptsize
\setlength{\tabcolsep}{7mm}{
\begin{tabular}{r|cc}\toprule
\multicolumn{1}{c|}{\textbf{Methods}} & \textbf{IoU (\%)} & \textbf{mIoU (\%)} \\\midrule
\textbf{Baseline (w/o KD)} & 44.19 & 10.56 \\
\textbf{Vanilla KD} & 44.51 &  11.27 \\
\textbf{MGD}~\pub{ECCV'22} & 43.29 & 11.26 \\
\textbf{DIST}~\pub{NeurIPS'22} & 43.11 & 11.40 \\
\textbf{CleanerS}~\pub{CVPR'23} & {{44.79}} & 11.86 \\
\textbf{RGO$^2$D (Ours)} & {43.80}& \textbf{13.27} \\
\bottomrule
\end{tabular}
}
\vspace{-2.5mm}
\caption{{Performance comparisons with other knowledge distillation schemes.}}
\label{tab:distill_comp}
\vspace{-4mm}
\end{table}

\begin{table}[!ht]\centering
\scriptsize
\setlength{\tabcolsep}{5.4mm}{
\begin{tabular}{r|ccc}\toprule
\multicolumn{1}{c|}{\textbf{Methods}} & \textbf{IoU (\%)} & \textbf{mIoU (\%)} \\\midrule
\textbf{RGO$^2$D \textit{w/o.} global-distill.} & 43.55 & 11.62 \\
\textbf{RGO$^2$D \textit{w/o.} local-distill.} & 43.46 & 12.86 \\
\textbf{RGO$^2$D \textit{w/o.} range-info.} & 43.56 & 12.78 \\
\textbf{RGO$^2$D \textit{w/o.} feature-distill.} & {44.77} & 12.67 \\
\multicolumn{1}{r|}{\textbf{RGO$^2$D}} & {43.80}& \textbf{13.27} \\
\bottomrule
\end{tabular}
}
\vspace{-2.5mm}
\caption{Ablation study for each item in our RGO$^2$D module.}
\label{tab:distill_scheme}
\vspace{-4mm}
\end{table}

\subsection{Ablation Studies}\label{sec:abl} 
In this section, we conduct ablation studies of our model components on the \textit{validation set} of SemanticKITTI.

\noindent \textbf{Effectiveness of Auto-Labelers.}
Firstly, we verify the effectiveness of scribbles as annotations to train the Dean-Labeler and Teacher-Labeler. 
As shown in Tab.~\ref{tab:scribble_ann}, the promising performance is achieved with our training pipeline compared with SOTA fully-supervised completion methods including LiDAR-based models. 
Our Dean-Labeler obtains 100\% IoU with the highest score of 42.28\% mIoU on ScribbleSC, which ensures that the quality of our pseudo-labels is sufficient to provide reliable supervision for the student model. The Teacher-Labeler also achieves a comparable performance with LiDAR-based methods and outperforms the camera-based models by a large margin. This observation reveals that the main bottleneck of current camera-based methods lies in the estimation of geometry.

\noindent \textbf{Impact of Each Module.}
Secondly, we study the impact of each module in the whole framework as shown in Tab.~\ref{tab:abl}. The Scribble2Scene with Dean-Labeler (DL) means that we replace the scribble annotations with the pseudo labels provided by DL. It can be seen that the models trained with DL obtain obviously higher geometric performance than those trained with scribbles directly. 
Moreover, we directly add the Teacher-Labeler (TL) with the Vanilla KD~\cite{hinton2015distilling}, and a slight precision improvement is achieved. Furthermore, we perform range-guided offline-to-online distillation (RGO$^2$D) with TL and the best performance is obtained.

\noindent \textbf{Effect of Offline-to-Online Distillation.}
Finally, we compare our RGO$^2$D  with other knowledge distillation methods including MGD~\cite{yang2022masked}, DIST~\cite{huang2022knowledge} for common KD and CleanerS~\cite{wang2023semantic} designed for SSC. As shown in Tab.~\ref{tab:distill_comp}, our proposed offline-to-online distillation scheme outperforms all other methods.
We further conduct the ablation analysis on each item of our distillation module as described in Tab.~\ref{tab:distill_scheme}. Compared to local distillation, the global one generates a larger performance impact.  
The range-guided information further enhances the performance by computing semantic logit and affinity at different ranges. With feature distillation, the online model obtains the best semantic scene completion accuracy. 

\section{Conclusion}
In this work, we have presented a scribble-based label-efficient benchmark ScribbleSC for semantic scene completion in autonomous driving. To enhance the performance in this setting, an effective scribble-supervised approach Scribble2Scene has been developed.
The offline Dean-Labeler provides dense semantic supervision and Teacher-Labeler guides the online model to learn structured occupancy information with new range-guided offline-to-online distillation. 
Extensive experiments demonstrate that our Scribble2Scene closes the gap between the sparse scribble-based approach and densely annotated methods, which shows competitive performance against the fully-supervised counterparts.

\section*{Acknowledgments}
This work is supported by National Natural Science Foundation of China under Grants (62376244). It is also supported by Information Technology Center and State Key Lab of CAD\&CG, Zhejiang University.

\bibliographystyle{named}
\bibliography{ijcai24}

\clearpage
\appendix
\section*{Supplementary Material}

\setcounter{figure}{0}
\setcounter{table}{0}
\renewcommand{\thefigure}{A\arabic{figure}}
\renewcommand{\thetable}{A\arabic{table}}

In this supplementary material, we extend our discourse to encompass the subsequent descriptions and experiments:

\begin{itemize}
    \item Section \ref{sec:supp_a}: More implementation details;
    \item Section \ref{sec:supp_b}: More details on ScribbleSC;
    \item Section \ref{sec:supp_c}: Additional results;
    \item Section \ref{sec:supp_d}: Limitations and our future work.

\end{itemize}

\section{More Implementation Details}
\label{sec:supp_a}
\noindent \textbf{Dean-Labeler.} As described in the main paper, Cylinder3D~\cite{zhu2021cylindrical} is adopted as the \textit{SCN} backbone. Since the occupancy space is divided by a cubic partition, we discard the cylindrical partition in Cylinder3D and adopt the same cubic partition.
The complete geometry $\mathcal{G}$ is converted into sparse tensor (\textit{Dense2Sparse}) with the same resolution as the occupancy space by \texttt{spconv}\footnote{https://github.com/traveller59/spconv.}. After extracting voxel-wise features by sparse convolution, we transform them into the original dense voxel space (\textit{Sparse2Dense}) and obtain the semantic occupancy predictions without loss of complete geometric information.

\noindent \textbf{Teacher-Labeler.} 
We adopt the 2$\times$ down-sampling complete geometry ${\mathcal{G}_{\frac{1}{2}}}$ as in VoxFormer~\cite{li2023voxformer} to perform deformable cross-attention~\cite{zhu2020deformable} with image features extracted by ResNet-50~\cite{he2016deep}. 
The image is cropped into the size of $1220 \times 370$. Besides the loss functions like partial cross-entropy loss $\mathcal{L}_\text{partial\_ce}$ and geometric scene-class affinity loss $\mathcal{L}_\text{geo}$, 
other training settings are the same as VoxFormer-T in order to facilitate fair comparison.

\noindent \textbf{Online Model.}
The network architecture of online model and training settings are similar to Teacher-Labeler, while we can only use the estimated noisy coarse geometry $\hat{\mathcal{G}_{\frac{1}{2}}}$ to sample image features. For the loss coefficients $w_r$ of different ranges in range-guided offline-to-online distillation, we set $r=1, 2, 3$ to $0.06$, $0.15$, and $0.2$ from near to far, which emphasizes the completion quality of the whole scene as our primary optimization target.
The trade-off weights $\alpha$, $\beta$ of the inter- and intra-relation loss in global distillation are set to $2.625$ and $0.375$, respectively.

\section{More Details on ScribbleSC}
\label{sec:supp_b}
\begin{figure*}
	\centering
	\subfigure[sequence 00]{
		\begin{minipage}[t]{0.5\linewidth}
			\centering
			\includegraphics[width=0.8\linewidth]{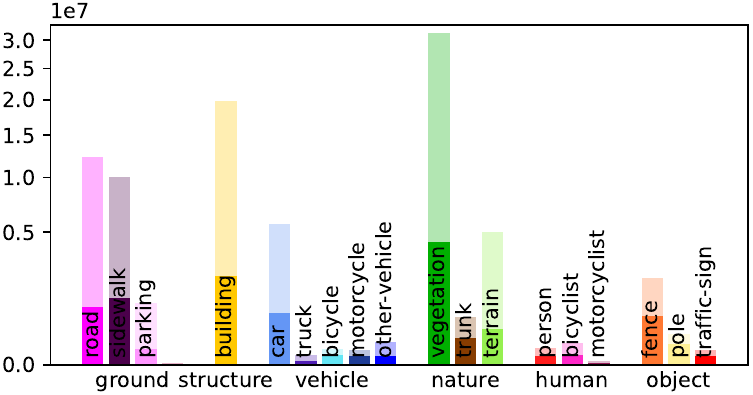}
		\end{minipage}
	}%
	\subfigure[sequence 01]{
		\begin{minipage}[t]{0.5\linewidth}
			\centering
			\includegraphics[width=0.8\linewidth]{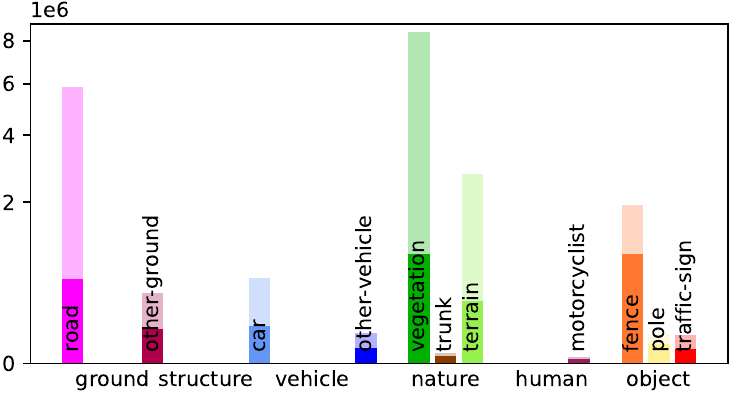}
		\end{minipage}
	}%

    \subfigure[sequence 02]{
		\begin{minipage}[t]{0.5\linewidth}
			\centering
			\includegraphics[width=0.8\linewidth]{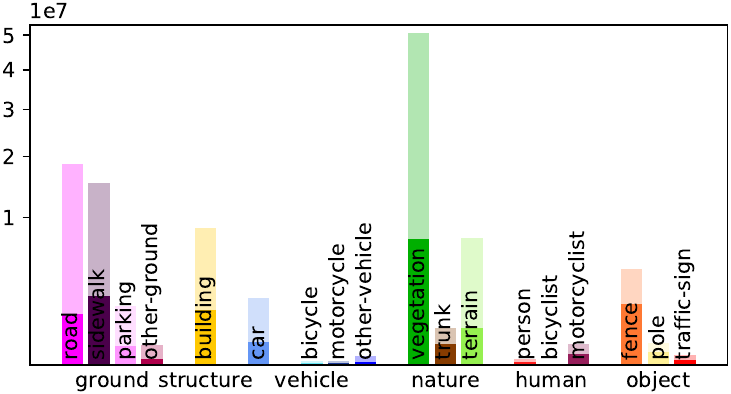}
		\end{minipage}
	}%
	\subfigure[sequence 03]{
		\begin{minipage}[t]{0.5\linewidth}
			\centering
			\includegraphics[width=0.8\linewidth]{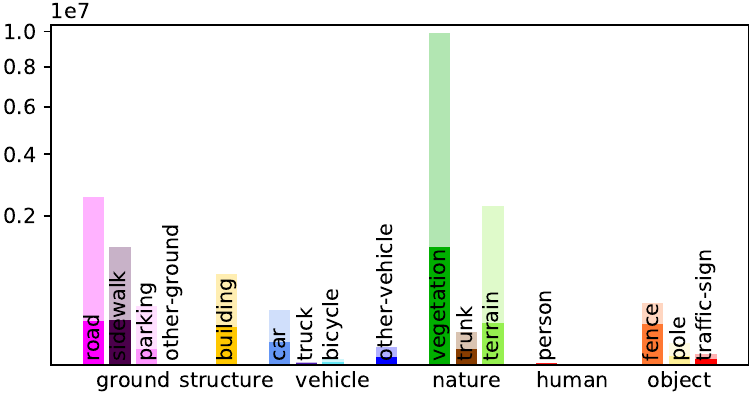}
		\end{minipage}
	}%

    \subfigure[sequence 04]{
		\begin{minipage}[t]{0.5\linewidth}
			\centering
			\includegraphics[width=0.8\linewidth]{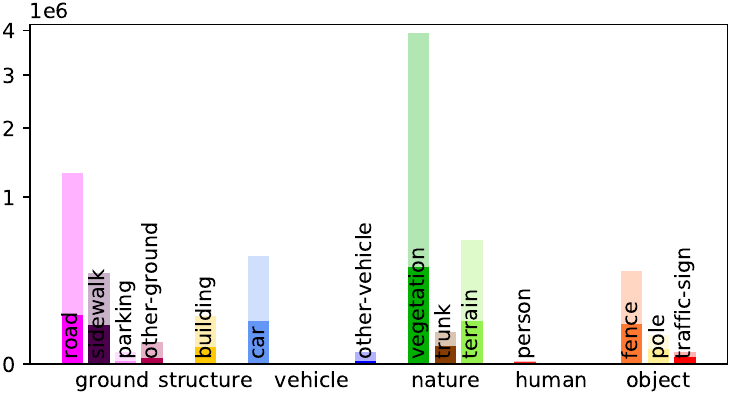}
		\end{minipage}
	}%
	\subfigure[sequence 05]{
		\begin{minipage}[t]{0.5\linewidth}
			\centering
			\includegraphics[width=0.8\linewidth]{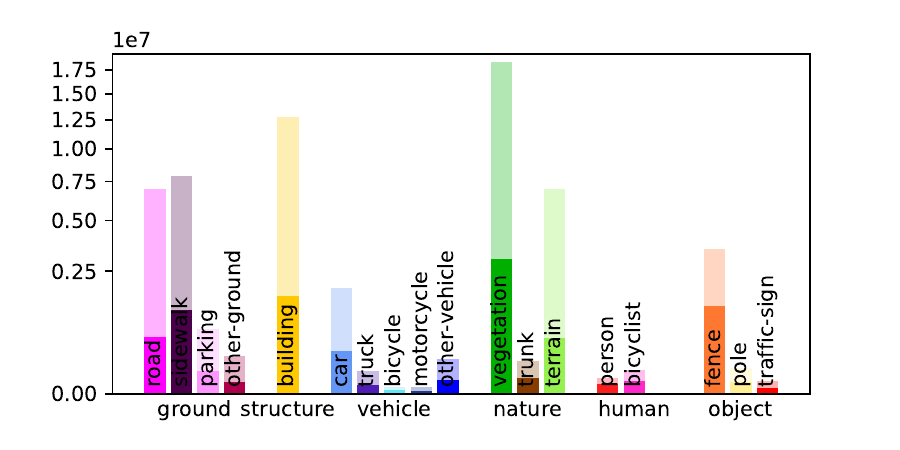}
		\end{minipage}
	}%

    \subfigure[sequence 06]{
		\begin{minipage}[t]{0.5\linewidth}
			\centering
			\includegraphics[width=0.8\linewidth]{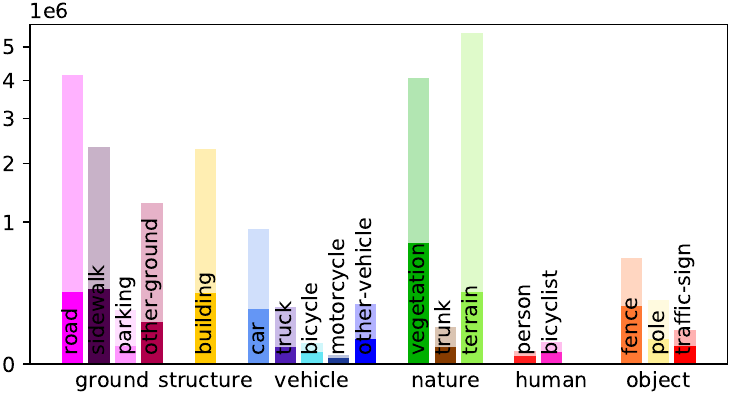}
		\end{minipage}
	}%
	\subfigure[sequence 07]{
		\begin{minipage}[t]{0.5\linewidth}
			\centering
			\includegraphics[width=0.8\linewidth]{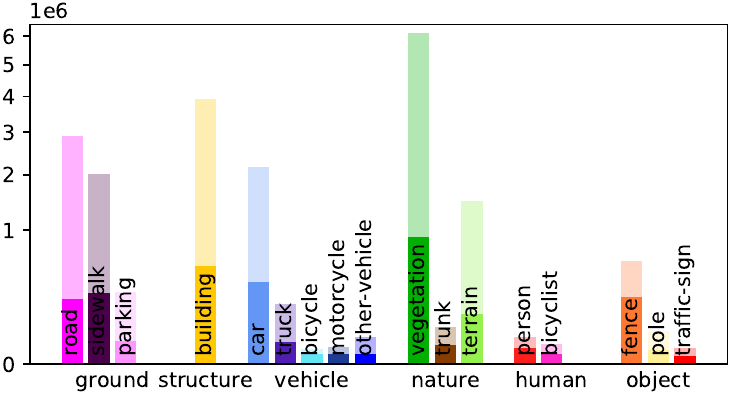}
		\end{minipage}
	}%

   \subfigure[sequence 09]{
		\begin{minipage}[t]{0.5\linewidth}
			\centering
			\includegraphics[width=0.8\linewidth]{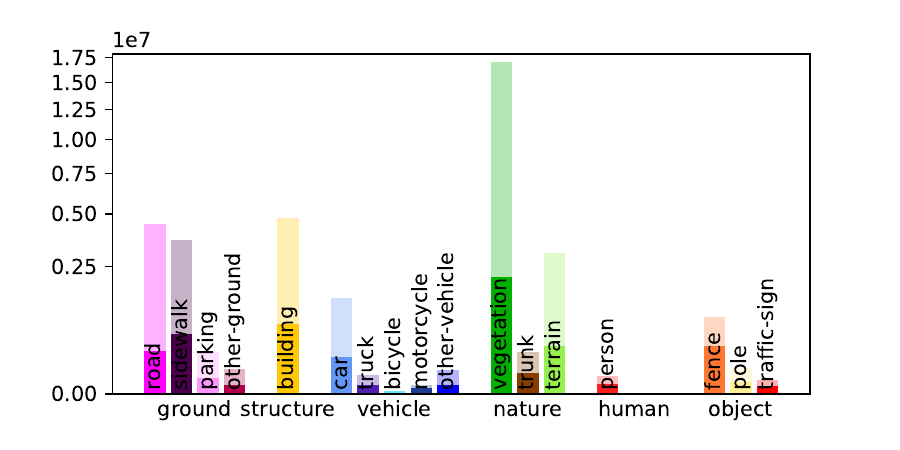}
		\end{minipage}
	}%
	\subfigure[sequence 10]{
		\begin{minipage}[t]{0.5\linewidth}
			\centering
			\includegraphics[width=0.8\linewidth]{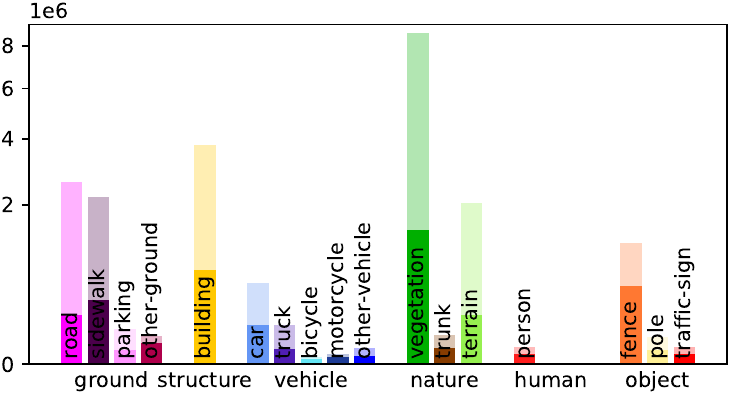}
		\end{minipage}
	}%

	\centering
	\caption{Quantitative results on each category of voxels labeled within ScribbleSC (deep color) comparing to the fully-annotated SemanticKITTI dataset (light color) in each sequence of the \textbf{\textit{training set}}.}
	\label{fig:supp_stat}
\end{figure*}

\noindent \textbf{Detailed Voxel Labeling Statistics.}
We provide the detailed voxel labeling statistics for each sequence in \textit{training set}. As illustrated in Fig.~\ref{fig:supp_stat}, both the fully-annotated SemanticKITTI SSC and scribble-annotated ScribbleSC have an obvious class-imbalance problem, where the background categories including \textit{vegetation} and \textit{road} account for the vast majority among these sequences. 
This challenge is particularly pronounced in the case of small objects such as \textit{bicycle}/\textit{bicyclist} and \textit{motorcycle}/\textit{motorcyclist} within ScribbleSC, which will guide our subsequent efforts in optimizing the labeling process and model training schemes.

\begin{figure*}[t]
\centering
\includegraphics[width=1.0\linewidth]{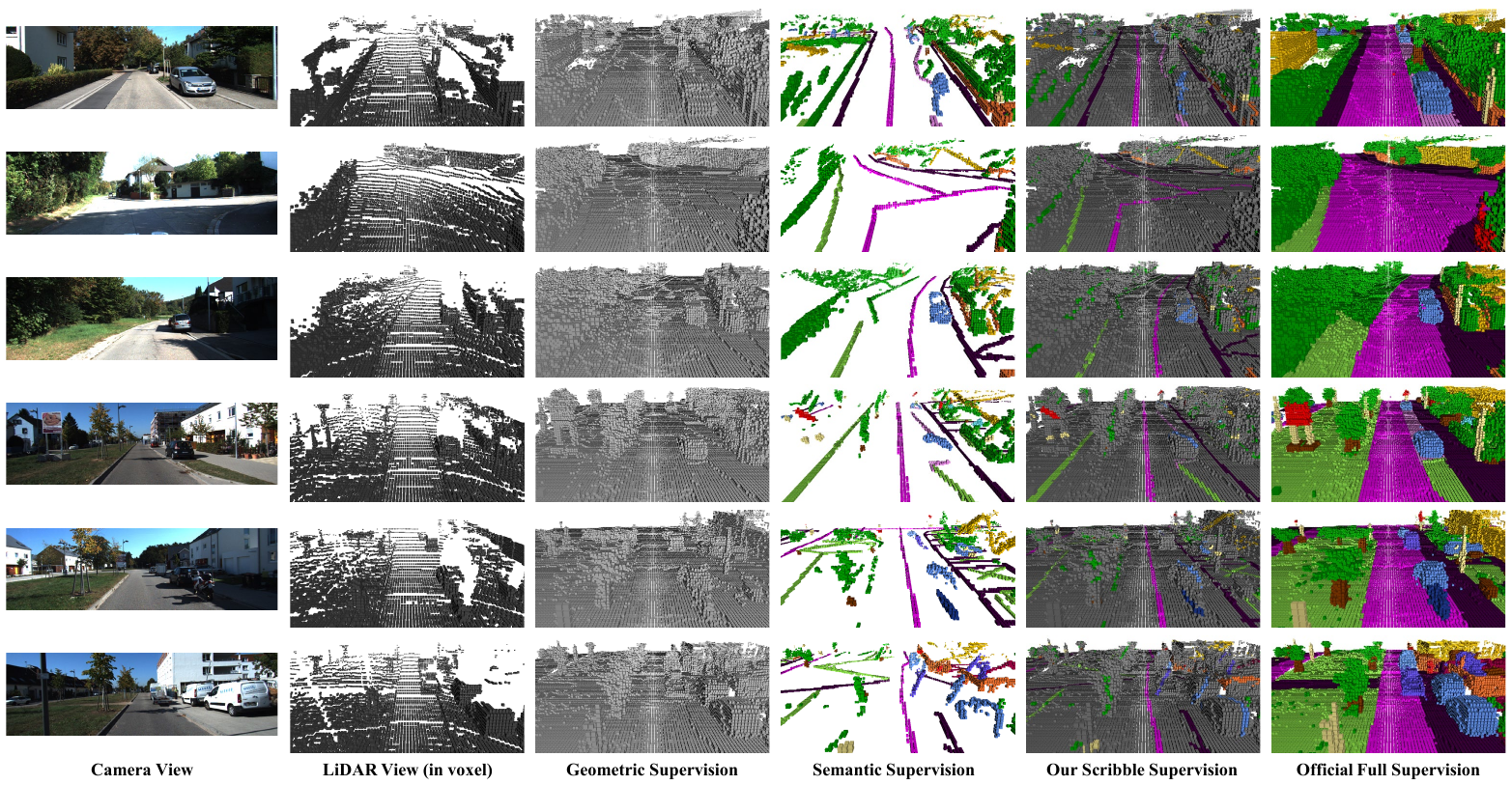}
    \caption{Qualitative visual comparisons of our constructed ScribbleSC with the scribble-based annotations and fully-annotated ground truth from SemanticKITTI. }
    \label{fig:supp_scri}
\end{figure*}

\noindent \textbf{Examples from ScribbleSC.}
More qualitative semantic scene completion examples from ScribbleSC are given in Fig.~\ref{fig:supp_scri}. Both the camera and LiDAR view (in voxel) are provided for better illustrations. The separate geometric and semantic supervisions are adopted to compute the geometric loss $\mathcal{L}_\text{geo}$ and partial cross-entropy loss $\mathcal{L}_\text{partial\_ce}$, respectively. In particular, the visual comparisons between our presented scribble supervision and the original full supervision are shown in the last two columns.

\section{Additional Results}
\label{sec:supp_c}
\subsection{Model Complexity Analysis}
We compare the methods with various model parameters, as shown in Tab.~\ref{tab:supp_complex}. Our online model has a similar number of parameters comparing to the baseline while achieving better performance with scribble annotations. The training time for online model is also close to the baseline.
During the inference, we only need to preserve the online model branch and obtain 1.05 FPS on a single GeForce RTX 4090 GPU as the fully-supervised model.
Our proposed online model training scheme does not rely on the specific model so that we can employ the models with different complexities and inference speeds as baselines.

\begin{table}[ht]\centering
\scriptsize
\renewcommand\arraystretch{0.75}
{
\begin{tabular}{r|c|cc}\toprule
\multicolumn{1}{c|}{\textbf{Methods}} & \textbf{Params.} & \textbf{IoU (\%)} & \textbf{mIoU (\%)}  \\\midrule
\textbf{VoxFormer (Baseline)}  & 57.9M & 37.76 & 10.42  \\
\textbf{Online Model (Ours)} & 58.2M & {43.80} & \textbf{13.27}  \\\midrule
\textbf{Dean-Labeler} & 58.3M & 100.00 & 42.28  \\
\textbf{Teacher-Labeler} & 57.9M & 82.80 & 21.70  \\
\bottomrule
\end{tabular}
}
\caption{Comparison of various model parameters in our approach.}
\label{tab:supp_complex}
\end{table}

\subsection{More Ablation Studies}
\noindent \textbf{Ablation on the location of distillation features.} To further examine the effectiveness of our proposed RGO$^2$D scheme, we conduct the ablation experiments on the location of the features for distillation.
As shown in Tab.~\ref{tab:distill_loc}, the first row denotes the results without feature-level distillation. 
Scribble2Scene achieves a promising performance with 13.27\% mIoU solely through the distillation of camera features. 
When we employ the features of \textit{DCA} or \textit{SSC Decoder} for distillation, the non-negligible performance degradation is observed. 
This phenomenon is partially attributed to the substantial geometric disparities between precise 
${\mathcal{G}_{\frac{1}{2}}}$
of Teacher-Labeler and estimated $\hat{\mathcal{G}_{\frac{1}{2}}}$ of online model.

\begin{table}[ht]\centering
\scriptsize
\setlength{\tabcolsep}{4.5mm}{
\begin{tabular}{lccccc}\toprule
\multicolumn{3}{c}{\textbf{Feature Location}} &\multirow{2}{*}{\textbf{IoU (\%)}} &\multirow{2}{*}{\textbf{mIoU (\%)}}  \\
\cmidrule(r){1-3}
A & B & C  & & \\ \midrule
& &   & 44.77 & 12.67 \\
\cmark & &  &   {43.80} &\textbf{13.27}     \\
&\cmark &  &  44.38 &  12.51  \\
& &\cmark &  43.85 &  12.36 \\
\cmark &\cmark &\cmark  & 43.74 & 12.82 \\
\bottomrule
\end{tabular}
}
\caption{{Ablation study on the location of the distillation features.} ``A'' denotes that the features originating from \textit{Camera Encoder}, ``B'' pertains to the features derived from \textit{DCA}, and ``C'' indicates the features arising  from \textit{SSC Decoder}. }
\label{tab:distill_loc}
\end{table}

\noindent \textbf{Ablation on the weight of distillation loss.}
Our overall loss function to train online model consists of semantic loss $\mathcal{L}_{\text{sem}}$, geometric loss $\mathcal{L}_{\text{geo}}$ and distillation loss $\mathcal{L}_{\text{distill}}$. We conduct ablation studies on the weight of $\mathcal{L}_{\text{distill}}$, as shown in Tab.~\ref{tab:distill_weight}. It can be seen that the proposed range-guided distillation scheme performs better with the appropriate weights.

\begin{table}[ht]\centering
\scriptsize
\renewcommand\arraystretch{0.75}
{
\begin{tabular}{cccccc}\toprule
\textbf{$\mathcal{L}_\text{distill}$ Weight} & 0.0 & 0.5 & 0.75 & 1.0 & 1.25  \\\midrule
\textbf{IoU (\%)} &  44.19 & 44.13 & 43.82 & 43.80 & 44.24 \\
\textbf{mIoU (\%)} & 10.56  & 12.59 & 12.75 & \textbf{13.27} & 12.96  \\
\bottomrule
\end{tabular}
}
\caption{Ablation study on the weight of range-guided offline-to-online distillation loss.}
\label{tab:distill_weight}
\end{table}

\subsection{Experiments on LiDAR-based Approaches}
\noindent \textbf{Implementation Details.} 
In the main paper, we provide quantitative comparisons with LiDAR-based methods on Semantic POSS~\cite{pan2020semanticposs}. 
For the baseline models including LMSCNet~\cite{roldao2020lmscnet} and MotionSC~\cite{wilson2022motionsc} with LiDAR input, the structure of Dean-Labeler is consistent with the camera-based implementation on SemanticKITTI~\cite{behley2019semantickitti}. 
We redesign the Teacher-Labeler with reference to the architecture of LMSCNet/MotionSC and replace the occupancy grid voxelized from the current LiDAR frame with the complete geometry $\mathcal{G}$.
The online model is the same as LMSCNet/MotionSC while being trained with pseudo labels provided by Dean-Labeler and offline-to-online distillation with Teacher-Labeler.

We present a detailed comparison of semantic categories on the \textit{validation set} of SemanticPOSS to supplement Tab.~3 in the main paper. As depicted in Tab.~\ref{tab:supp_poss}, our method achieves competitive performance across most classes comparing to those fully-supervised methods.

\begin{table*}[t]
\renewcommand\tabcolsep{4pt}
\scriptsize
\newcommand{\classfreq}[1]{{~\scriptsize(\semkitfreq{#1}\%)}} 
    \centering
    \begin{tabular}{r|c|c|c|c|c|c|c|c|c|c|c|c|c|c|c}
   \toprule 
\multicolumn{1}{c|}{\textbf{Methods}}
&\rotatebox{90}{\textbf{Supervision}}
&\rotatebox{90}{\textbf{IoU (\%)}}
& \rotatebox{90}{\textbf{person}}
& \rotatebox{90}{\textbf{rider}}
& \rotatebox{90}{\textbf{car}} 
& \rotatebox{90}{\textbf{trunk}} 
& \rotatebox{90}{\textbf{plants}}
& \rotatebox{90}{\textbf{traf.-sign}}
& \rotatebox{90}{\textbf{pole}}
& \rotatebox{90}{\textbf{building}}
& \rotatebox{90}{\textbf{fence}}
& \rotatebox{90}{\textbf{bike}} 
& \rotatebox{90}{\textbf{ground}}

&\rotatebox{90}{\textbf{mIoU (\%)}}
&\rotatebox{90}{\textbf{SS/FS (\%)}}
\\ \midrule

\textbf{LMSCNet}~\pub{3DV'20} & Fully & 54.27 & 5.00 & 0.00 & 0.64  & 2.31 & 40.34 & 2.96  & 2.51  &  36.65  & 11.93  & 29.27  & 47.05   & 16.24 & - \\
\textbf{LMSCNet}$^{\dag}$~\pub{3DV'20} & Sparse & 31.00 & \tbblue{11.08} & \tbblue{0.10} & 0.40  &  3.00 & 23.38 & 0.81  & \tbblue{10.76}  &  23.05  &  9.10 &  19.63 &   33.54 & 12.26 & 75.49 \\

\textbf{LMSCNet with S2S (Ours)} &  Sparse & \tbblue{53.24} & 7.25  & 0.00 & \tbblue{0.58}  & \tbblue{3.43} & \tbblue{39.09} &  \tbblue{2.31} & 1.02  & \tbblue{37.72}   &  \tbblue{12.91} &  \tbblue{28.63} &  \tbblue{43.22}  & \tbblue{16.01} & \tbblue{98.58}   \\\midrule
\textbf{MotionSC}~\pub{RA-L'22} & Fully & 53.28 & 4.11  & 0.00 &  1.72 & 3.84 & 40.14 & 3.22  &  6.40 & 37.53 & 14.84  &  34.82 &  52.54  & 18.10 & -  \\
\textbf{MotionSC}$^{\dag}$~\pub{RA-L'22} & Sparse & 33.45  & \tbblue{9.78} & \tbblue{0.44}  & 0.77 & 2.55 & 26.47  & \tbblue{5.71}  &  \tbblue{4.00}  & 31.11  & \tbblue{13.72}  &  18.06  &  34.51 & 12.91 & 71.33\\

\textbf{MotionSC with S2S (Ours)} &  Sparse & \tbblue{53.48} & 5.56 & 0.00 &  \tbblue{2.53} & \tbblue{3.79} & \tbblue{41.06} & 2.30  &  3.99 &  \tbblue{37.76}  & 12.05  &  \tbblue{33.87} & \tbblue{50.99}   &  \tbblue{17.63} & \tbblue{97.40} \\

\bottomrule
\end{tabular}
    \caption{Quantitative comparisons with the LiDAR-based baseline models on the \textit{\textbf{{validation set}}} of SemanticPOSS. $\dag$ indicates the results that are directly trained with the sparse semantic labels. The best results in scribble-supervised models are marked in \tbblue{blue}.}
    \label{tab:supp_poss}
\end{table*}

\noindent \textbf{More Experiments on SemanticKITTI.}
We also conduct experiments with LiDAR input on SemanticKITTI~\cite{behley2019semantickitti} to demonstrate the scalability of Scribble2Scene (S2S). 
As illustrated in Tab.~\ref{tab:supp_lidar}, our method achieves significant performance improvement (15.38\% mIoU \textit{v.s.} 10.03\% mIoU, 16.46\% mIoU \textit{v.s.} 11.18\% mIoU) compared with these models trained with ScribbleSC directly.

\begin{table}[ht]\centering
\scriptsize
\setlength{\tabcolsep}{1.4mm}{
\begin{tabular}{r|c|cc|c}\toprule
\multicolumn{1}{c|}{\textbf{Methods}} & \textbf{Supervision} & \textbf{IoU (\%)} & \textbf{mIoU (\%)} & \textbf{SS/FS (\%)} \\\midrule
\textbf{LMSCNet}~\pub{3DV'20} & Fully & 53.92 & 16.12 & - \\
\textbf{LMSCNet$^{\dag}$}~\pub{3DV'20} & Scribble & 31.54 & 10.03 & 62.22 \\
\textbf{LMSCNet with S2S (Ours)} & Scribble & 54.05 & \textbf{15.38} & \textbf{95.41} \\\midrule
\textbf{MotionSC}~\pub{RA-L'22} & Fully & 54.72 & 16.61 & - \\
\textbf{MotionSC$^{\dag}$}~\pub{RA-L'22} & Scribble & 39.08 & 11.18 & 67.31 \\
\textbf{MotionSC with S2S (Ours)} & Scribble & {55.87}& \textbf{16.46} & \textbf{99.10} \\
\bottomrule
\end{tabular}
}
\vspace{-2mm}
\caption{Quantitative results with LiDAR input on the \textit{\textbf{validation set}} of SemanticKITTI. $\dag$ represents the model that is directly retrained with ScribbleSC.}
\label{tab:supp_lidar}
\vspace{-2mm}
\end{table}

\subsection{Qualitative Comparison}
\begin{figure*}[t]
\centering
\includegraphics[width=1.0\linewidth]{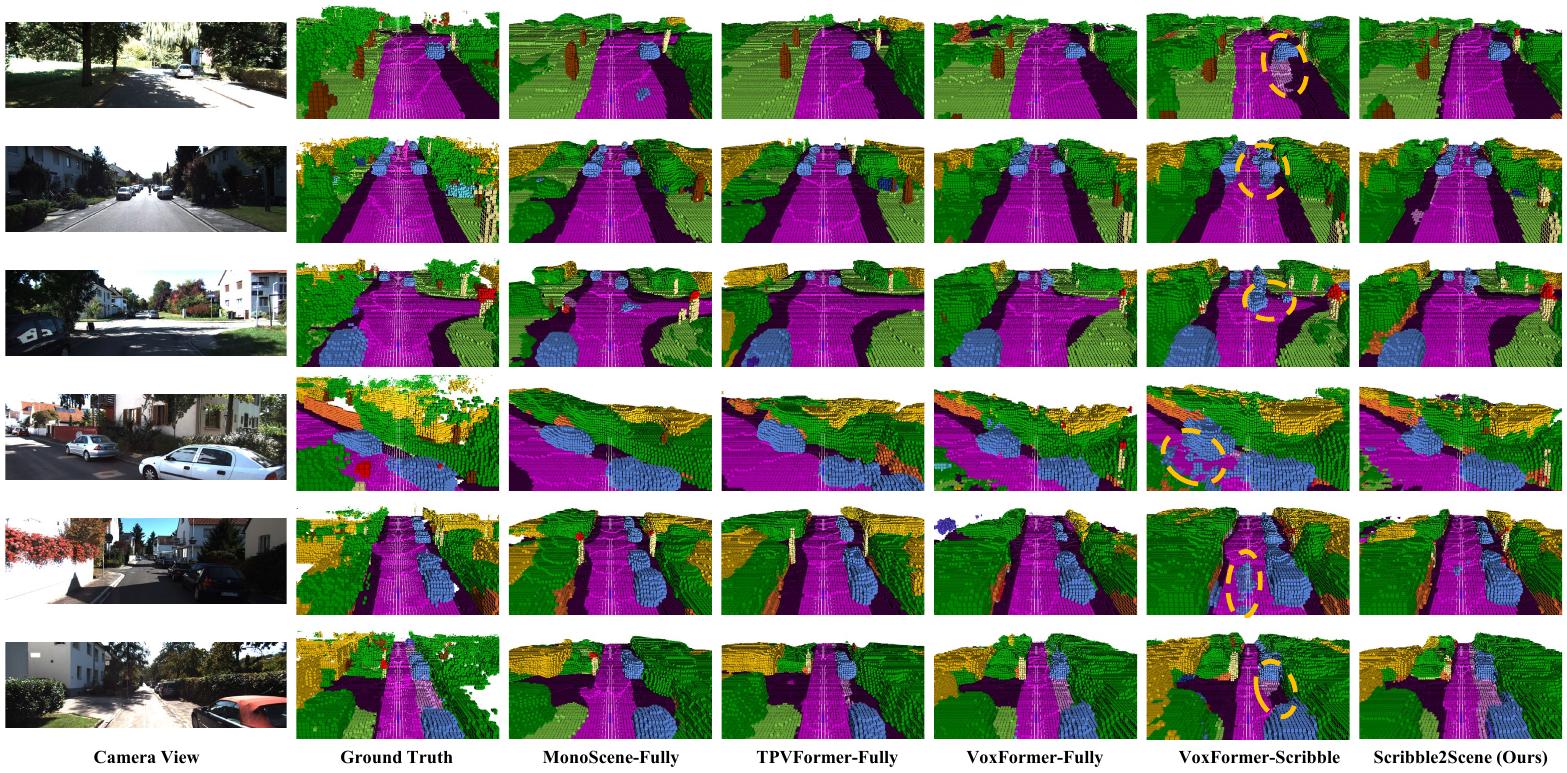}
    \caption{Qualitative comparisons against the  camera-based SSC methods with fully supervision and baseline model on the \textit{\textbf{validation set}} of SemanticKITTI. We compare our proposed Scribble2Scene approach with the state-of-the-art camera-based methods including MonoScene, TPVFormer and VoxFormer.}
    \label{fig:supp_viscomp}
\end{figure*}

Fig.~\ref{fig:supp_viscomp} reports the additional visual results on the \textit{validation set} of SemanticKITTI.
It can be seen that Scribble2Scene obtains the stable performance compared with the baseline model and even state-of-the art fully-supervised methods~\cite{cao2022monoscene,huang2023tri,li2023voxformer}, particularly in the semantic categories such as \textit{car} and \textit{sidewalk} which hold significance for the development of autonomous driving systems.

\section{Limitations and Our Future Work}
\label{sec:supp_d}
Due to the limited occurrences, our presented Scribble2Scene method may not perform well on objects with long-tailed distributions,  especially on small objects with sparse annotations. 
This is a typical problem for both scribble-supervised and fully-supervised methods as discussed in Sec.~\ref{sec:supp_b}. We will explore the effective data augmentation methods to expand the samples, like the copy-paste scheme commonly used in 2D space. Moreover, generating training samples for these hard classes through simulators (\textit{e.g.}, CARLA) can further contribute to the upcoming research.

Aside from the above challenges, we construct the sparse semantic labels and conduct the experiments on SemanticKITTI also SemanticPOSS, which still entails a certain amount of human efforts to provide the initial scribble/sparse annotations. In the future, we will explore the approach in a fully label-free manner based on foundation models, such as large vision-language models.

\end{document}